\documentclass[review,3p,times]{elsarticle}

\usepackage{amssymb}
\usepackage{latexsym}
\usepackage{graphicx}
\usepackage{amsfonts,amssymb,amsmath}
\usepackage{hyperref}
\usepackage{adjustbox}
\usepackage{array}
\usepackage{multirow,booktabs}
\usepackage[utf8]{inputenc}
\usepackage[english]{babel}
\usepackage{xcolor}
\usepackage{caption} 
\usepackage{ulem}
\usepackage{subcaption}

\begin{document}
\newcommand{\noteBK}[1
]{\textcolor{purple}{[{\bf #1}]}}

\begin{frontmatter}



\title{Rethinking Detection Based Table Structure Recognition for Visually Rich Document Images}

\author[inst1]{Bin Xiao}
\ead{bxiao103@uottawa.ca}
\author[inst1]{Murat Simsek}
\ead{murat.simsek@uottawa.ca}
\author[inst1]{Burak Kantarci}
\ead{burak.kantarci@uottawa.ca}

\affiliation[inst1]{organization={School of Electrical Engineering and Computer Science},
            addressline={University of Ottawa}, 
            city={Ottawa},
            postcode={K1N 6N5}, 
            state={ON},
            country={Canada}}

\author[inst2]{Ala Abu Alkheir}
\ead{ala\_abualkheir@lytica.com}

\affiliation[inst2]{organization={Lytica Inc.},
             addressline={555 Legget Drive}, 
            city={Ottawa},
            postcode={K2K 2X3}, 
            state={ON},
            country={Canada}}

\begin{abstract}
Table Structure Recognition (TSR) is a widely discussed task aiming at transforming unstructured table images into structured formats, such as HTML sequences, to make text-only models, such as ChatGPT, that can further process these tables. One type of solution is using detection models to detect table components, such as columns and rows, then applying a rule-based post-processing method to convert detection results into HTML sequences. However, existing detection-based models usually cannot perform as well as other types of solutions regarding cell-level TSR metrics, such as TEDS, and the underlying reasons limiting the performance of these models on the TSR task are also not well-explored. Therefore, we revisit existing detection-based models comprehensively and explore the underlying reasons hindering these models' performance, including the improper problem definition, the mismatch issue of detection and TSR metrics, the characteristics of detection models, and the impact of local and long-range features extraction. Based on our analysis and findings, we apply simple methods to tailor a typical two-stage detection model, Cascade R-CNN, for the TSR task. The experimental results show that the tailored Cascade R-CNN based model can improve the base Cascade R-CNN model by 16.35\% on the FinTabNet dataset regarding the structure-only TEDS, outperforming other types of state-of-the-art methods, demonstrating that our findings can be a guideline for improving detection-based TSR models and that a purely detection-based solution is competitive with other types of solutions, such as graph-based and image-to-sequence solutions. 

\end{abstract}

\begin{highlights}

\item Revealing the issues of detection-based Table Structure Recognition (TSR) studies.

\item Analysis and explanation of critical aspects of detection models and the TSR task.

\item Proposing a Cascade R-CNN based solution, based on the analysis.

\item Achieving STOA performance on various datasets regarding COCO and TEDS metrics. 

\item Further explanation on the impact of critical aspects based on experimental results.

\end{highlights}


\begin{keyword}
Table Structure Recognition \sep Information Extraction \sep Document Processing  \sep Object Detection \sep Visually Rich Document Understanding
\end{keyword}

\end{frontmatter}


\section{Introduction}
\label{sec:introduction}



Datasheets and scanned documents are widely used in business scenarios and on the internet because of their user-friendliness to human readers. However, these documents are usually not structured, which means that it is hard to extract useful information from them and further analyze the semantic meaning automatically. Besides the unstructured format of the document, tables in these documents, which are widely used to summarize critical information, often have very complex structures and layouts, making it challenging to interpret and analyze these tables. Therefore, there have been many studies~\cite{rastan2019texus, yildiz2005pdf2table, mendes2017tabula, fernandes2023tablestrrec, xiao2022efficient, xiao2022handling, zhong2020image, zheng2021global, qiao2021lgpma,ma2023robust,li2022table,huang2023improving,nassar2022tableformer,visapp23namly} discussing the Table Structure Recognition (TSR) problem aiming at transforming unstructured table images into structured formats, such as HTML sequences, so that the transformed structured tabular data can be further analyzed by text-only models, such as ChatGPT~\cite{openai_chatgpt}. Even though some studies and tools~\cite{rastan2019texus, yildiz2005pdf2table, mendes2017tabula} can parse the Portable Document Format (PDF) files directly to extract the content and tables, these solutions cannot deal with scanned documents and table images and their performance is often limited on the TSR problem. Therefore, this study focuses on the deep learning based models.

TSR studies can roughly be categorized into three groups based on their problem formulations: detection-based models, image-to-sequence models, and graph-based models. Image-to-sequence models usually follow the encoder-decoder architecture and directly generate structured outputs, such as HTML sequences. Some image-to-sequence models~\cite{visapp23namly, ye2021pingan} also integrate the OCR task into the model to make the model end-to-end without using extra OCR tools~\cite{mmocr2021,xiao2023multi} to extract text contents from the images. However, since these models use auto-regressive decoders, they often suffer from error accumulation~\cite{shen2023divide}, and their OCR capacity usually cannot generalize well because of the limitation of training data. On the other hand, graph-based models usually use segmentation or detection methods to extract table cells, treat extract table cells as nodes of a graph, and further build the relation among the graph nodes. This graph-based definition makes it easier to deal with the scenarios in which table images are collected from the wild, such as rotated, distorted tables. However, graph-based models introduce extra complexity because they need to build extra graph models compared with detection-based models. By contrast, detection-based models are more straightforward in detecting the table components directly and post-processing the detection results with a deterministic rule-based method for reconstructing the table structure. However, detection-based methods can fail to deal with rotated and distorted samples. Besides, detection-based models usually cannot perform as well as other types of solutions regarding cell-level TSR metrics, such as TEDS~\cite{zhong2020image}. Therefore, these different types of approaches have their benefits and must be selected based on the application scenarios. In this study, we focus on the table images from well-formatted documents and explore the underlying reasons hindering the performance of detection-based models.

There have been many studies~\cite{siddiqui2019deeptabstr,hashmi2021guided,fernandes2023tablestrrec,xiao2022efficient,smock2022pubtables} using detection models together with a post-processing method to solve the TSR task. However, existing studies either over-simplify the problem or define a multi-label detection problem, which is challenging for two-stage object detectors. For example, some studies~\cite{siddiqui2019deeptabstr,hashmi2021guided, fernandes2023tablestrrec, xiao2022efficient} do not define Column Header as detection target, making it impossible to provide information regarding the header cells. By contrast, PubTables1M~\cite{smock2022pubtables} defines six types of components, including Table, Column, Row, Spanning cell, Column Header, and Projected Row Header, which can provide as much structure information as other types of TSR models. However, PubTables1M~\cite{smock2022pubtables} does not consider that some Column Headers and Projected Row Headers can share identical bounding boxes with corresponding Rows, making this definition a multi-label detection task. Besides these issues of problem formulation, detection models used in these studies are trained to optimize their detection performance. However, since the complex structures of tables are processed by a post-processing step inferring defined table components, such as Columns and Rows, a model with good detection performance cannot necessarily lead to good performance in TSR metrics, such as TEDS. Moreover, some critical characteristics of detection models are not considered in the model design and problem formulation in existing studies. For example, typical two-stage detection models, such as Faster R-CNN~\cite{ren2015faster} and Cascade R-CNN~\cite{Cai2018CascadeRD}, are not suitable for multi-label detection tasks, while transformer-based detection models, such as DETR~\cite{carion2020end} and Sparse R-CNN~\cite{sun2021sparse}, can achieve promising results on multi-label detection tasks. Another example is that for two-stage detection models, regional proposal generation plays a crucial role in the model's performance, and the defined components in table images have different aspect ratios compared with common objects. At last, many studies apply deformable convolution~\cite{dai2017deformable} to improve the models' performance regarding detection evaluation metrics, such as COCO metric~\cite{lin2014microsoft}. However, simply applying deformable convolution can degrade the model's performance regarding the TEDS, and it is necessary to extract long-range dependencies while improving the local feature extraction. Therefore, in this study, we comprehensively revisit existing detection-based solutions and further explore the possible reasons hindering the performance of detection-based models for the TSR task. Based on our findings and analysis, we apply three simple methods to a typical two-stage detection model, Cascade R-CNN, including tuning the aspect ratios and increasing the number of region proposals in regional proposal generation, transforming the multi-label detection task into the single-label task, and introducing a Spatial Attention Module to build long-range dependencies. The experimental results show that our proposed method can achieve state-of-the-art performance with very simple methods, demonstrating that our findings can be a guideline for further improvement of detection-based solutions. To sum up, the contributions of this study are four-fold: 

\begin{enumerate}

\item We comprehensively revisit existing detection-based TSR models and explore possible reasons hindering the performance of these models, including the problem formulation, the mismatch of optimization target of detection models and the TSR evaluation metrics, the characteristics of detection models, and the impact of feature extraction. Our analysis and findings can be a guideline for further improving the performance of detection-based TSR models.

\item Based on our analysis and findings, we apply three simple methods to improve Cascade R-CNN, including proposing a pseudo-class generation method to transform multi-label detection into a regular single-label detection problem, adjusting the ratio aspects and the number of regional proposals in the region proposal generation, applying the deformable convolution and introducing a Spatial Attention Module to build the long-range dependencies and context information in the backbone network.

\item We conduct extensive experiments to evaluate our proposed solution on various datasets, including SciTSR~\cite{chi2019complicated}, FinTabNet~\cite{zheng2020global}, PubTabNet~\cite{zhong2020image} and PubTables1M~\cite{smock2022pubtables} with both detection metrics and cell-level TSR metrics. The experimental results show that our proposed solution can outperform state-of-the-art models in terms of detection and cell-level TSR metrics. 

\item We further verify our analysis and findings with experiments and discuss and explain more useful insights from the experimental results for further model design. 

\end{enumerate}
The rest of this paper is organized as follows: Section~\ref{sec:related_work} discusses related studies, including studies in Object Detection and Table Structure Recognition. Section~\ref{sec:rethinking_detection_based_models} explores and discusses the reasons that hinder the performance of detection-based TSR models. Section~\ref{sec:proposed_method} describes our proposed solution based on our analysis and findings. Section~\ref{sec:experiments} shows the experimental results and discusses the design aspects of the proposed method. At last, we draw our conclusion and possible directions in section~\ref{sec:conclusion}.


\section{Related Work}
\label{sec:related_work}

\subsection{Object Detection Models}
Object Detection is a fundamental task that has been widely discussed in many studies. Since deep models have become the dominant solutions, we only discuss popular deep learning based models in this section. Based on different design perspectives, popular detection models can be categorized in different ways. One popular categorization of detectors based on the number of regression steps is to classify them into one- and two-stage detectors. Two-stage models, such as Faster R-CNN~\cite{ren2015faster} and Cascade R-CNN~\cite{Cai2018CascadeRD}, usually use a Region Proposal Network (RPN) to generate region proposals first and then feed the region proposals to the well-designed model to classify and regress the proposals. In the RPN network, one key parameter is the aspect ratio, which defines the height/width ratio when generating anchor boxes. Suitable aspect ratios are often close to the target objects' height/width ratio, making the regression task easier and improving the model performance. In contrast, popular one-stage models, such as FCOS~\cite{tian2019fcos}, YOLO~\cite{redmon2016you}, and YOLO's variants~\cite{redmon2017yolo9000, redmon2018yolov3, bochkovskiy2020yolov4}, integrate the region proposal generation and other regression and classification components into a single network. For example, YOLO divides the images into grids first, then classifies the class of grid cells and directly predicts the bounding boxes and their confidences. The simple design of one-stage detectors leads to faster training and inference time compared with two-stage detectors.

On the other hand, some studies~\cite{sun2021sparse, zhang2023dense} categorize the popular detectors from the perspective of Non-maximum Suppression (NMS), which is widely used to reduce the redundant predictions from the detectors. From this perspective, popular detectors can be categorized into end-to-end and none end-to-end models based on whether NMS is needed. DETR~\cite{carion2020end} is a typical end-to-end detector introducing transformer architecture~\cite{vaswani2017attention}, set prediction loss, and one-to-one label assignment to the object detection problem. Sparse R-CNN~\cite{sun2021sparse} further refactors the DETR model and proposes to use sparse learnable regional proposals to replace dense regional proposals and utilize a dynamic instance interactive head to regress and classify the proposals in an iterative manner. Study~\cite{sun2021makes} analyses the success of end-to-end detectors and argues that the one-to-one label assignment method in end-to-end detectors contributes to the success of the end-to-end models but is not sufficient to fully remove the NMS from the pipeline. This study further points out that the classification cost in the matching cost when applying one-to-one label assignment plays a key role in the success of these end-to-end models. Study~\cite{zhang2023dense} further analyzes combinations of the label assignment methods and queries and argues that sparse queries with one-to-one label assignment can degrade the recall, and dense queries with one-to-one label assignment are hard to optimize. To address these issues, study~\cite{zhang2023dense} proposes a dense distinct queries (DDQ) method to select distinct queries from dense queries using a class-agnostic NMS, achieving promising precision and recall. It is worth mentioning that these end-to-end detector can easily be extended to none end-to-end solutions by adapting many-to-one label assignments~\cite{hong2022dynamic} and NMS. In this study, we refer to models using transformer architecture, set prediction loss, and their variations as transformer-based detection models, such as DETR, Sparse R-CNN and Deformable-DETR~\cite{zhu2020deformable}.

\subsection{Table Structure Recognition}
There have been many studies~\cite{Chi2019ComplicatedTS, Adiga2019TableSR, zheng2021global, xiao2022efficient, Schreiber2017DeepDeSRTDL, liu2022neural} discussing the TSR problem in recent years. As mentioned earlier in Section~\ref{sec:introduction}, we can roughly categorize these solutions into image-to-sequence, detection-based, and graph-based models. Image-to-sequence based models usually define the ground truth as structured sequences, such as HTML sequences, built on the transformer architecture~\cite{vaswani2017attention}, and follow an encoder-decoder architecture. For instance, TableMaster~\cite{ye2021pingan} is a typical image-to-sequence based model that can generate HTML sequences. More specifically, TableMaster follows the architecture of MASTER~\cite{lu2021master}, which is originally designed for the scene text generation following the transformer architecture~\cite{vaswani2017attention}, and further improved the encoder part by introducing a Multi-Aspect Global Context Attention. Besides, TableMaster has two branches designed for the HTML sequence generation and bounding box regression. Similarly, MTL-TabNet~\cite{visapp23namly} also follows the encoder-decoder architecture but contains three decoders for the cell box regression, cell content recognition, and HTML sequence generation, respectively. DRCC~\cite{shen2023divide} argues that the error accumulation problem degrades the performance of image-to-sequence TSR models, especially when the input image is large. Therefore, DRCC proposes a two-step decoder architecture, which first decodes the input image into rows and then decodes the rows in cell sequences. VAST~\cite{huang2023improving} pays more attention to the imprecise bounding boxes of table cells and proposes a Coordinate Sequence Decoder to improve the model's ability to generate accurate bounding boxes and introduces a visual-alignment loss to align the visual and structural information. To sum up, this type of method is usually based on the encoder-decoder architecture and can be trained end-to-end without using post-processing methods. Since the ground truth sequences used in image-to-sequence models usually contain information regarding spanning cells and header cells, these models can handle complex structures with spanning cells and identify header cells.

On the other hand, detection-based models usually define the problem as detecting different table components and applying a post-processing method to reconstruct table structures. DeepTabStR~\cite{siddiqui2019deeptabstr} proposes to detect columns and rows to obtain the table cells. However, DeepTabStR ignores the row/column-span in the tables, which means that it cannot recover the hierarchical structures of tables. TableStrRec~\cite{Fernandes2023TableStrRecFF} extends the DeepTabStR, defining four types of table components: regular columns, irregular columns, regular rows, and irregular rows. Then, the spanning cells across multiple columns can be inferred from the difference between the regular and irregular columns when they are overlapped, and the spanning cells across multiple rows can be inferred from regular and irregular rows similarly. PubTables1M~\cite{smock2022pubtables} is another typical detection-based approach that defines six table components: table, column, row, spanning cell, Projected Row Header, and Column Header, in which Projected Row Header and Column Header are for the function analysis, and other components can be used to reconstruct the complex table structure. Among these formulations, only the problem formulation of PubTables1M can provide as much information as image-to-sequence models because it can provide header cell information and reconstruct the complex table structure. Besides, these detection-based models need an extra deterministic rule-based post-processing method to infer the table structure from detected table components, meaning they are not end-to-end. 

At last, graph-based methods usually apply either detection or segmentation methods to obtain the locations of table cells and further build the relation among table cells. For instance, TGRNet~\cite{xue2021tgrnet} formulates the cell location detection and cell logical location prediction jointly in a multi-task architecture, which is modularized by a segmentation based method and graph convolutional network (GCN), respectively. Similarly, TSRNet~\cite{li2022table} proposes a unified GNN-based approach modeling table detection and table structure recognition tasks together. More specifically, TSRNet also employs a semantic segmentation module to extract primitive regions, then applies k-nearest neighbors and line-of-sight neighbors to construct the graph and further classify the graph nodes and edges to filter the noise regions, merge, and build relations. In contrast, LGPMA~\cite{qiao2021lgpma} proposes a Local Pyramid Mask Alignment Module and Global Pyramid Mask Alignment Module to localize table cells, which are formulated as detection and segmentation problems and can be implemented by MaskR-CNN~\cite{he2017mask}. To construct the structure of the table, LGPMA further proposes a pipeline of cell matching, empty cell searching, and empty cell merging using the Maximum Clique Search algorithm and rule-based methods. Besides building graph explicitly, some studies~\cite{tensmeyer2019deep, zhang2022split, nguyen2023formerge} predict the table grids or separators first, and then merge grid elements, which are also treating grid elements as graph nodes. SPLERGE~\cite{tensmeyer2019deep} is a typical method following this strategy consisting of a Split Model and Merge Model, in which the Split Model consists of a Row Projection Network and a Columns Projection Network to obtain the table grid, and the Merge Model is used to merge the grid cells. Similarly, SEM~\cite{zhang2022split} employs a segmentation model to segment columns and rows and generate the table grid with a post-processing method. After the table grid is obtained, SEM introduces an Embedder network to extract and fuse the features from textual and visual modalities. A Merger network takes the fused features from Embedder as inputs to merge the grid elements. TSRFormer-DQ-DETR~\cite{WANG2023109817} leverages a DETR ~\cite{carion2020end} based separation line prediction model, termed DQ-DETR, to predict the reference points on separation lines, followed by a Relation Network based cell Merging module to merge grid elements. Since these graph-based models identify the graph nodes first, defining a cell-type classification task is necessary if they want to provide information regarding header cells.

\section{Rethinking Detection-based TSR models}
\label{sec:rethinking_detection_based_models}
\subsection{Preliminaries}
\label{sec:prelimiaries}
Since most existing detection-based TSR models are based on two-stage and transformer-based detectors, we use Cascade R-CNN~\cite{Cai2018CascadeRD} and Sparse R-CNN~\cite{sun2021sparse} as two examples of these two types of detectors and briefly review their critical designs in this section. 

\subsubsection{Cascade R-CNN}
\label{sec:cascade_r_cnn}
\begin{figure*}[htp]
\begin{center}
\includegraphics[width=\textwidth, keepaspectratio]{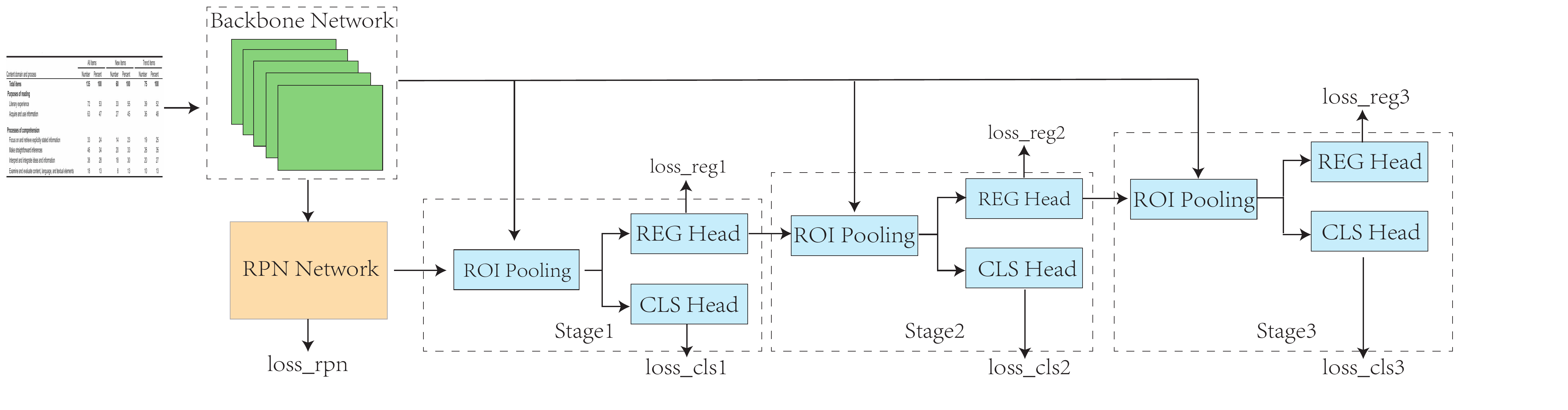}
\caption{Overall architecture of Cascade R-CNN.}
\label{fig:overall_architecture_cascade_rcnn}
\end{center}
\end{figure*}

Cascade R-CNN~\cite{Cai2018CascadeRD} is a typical two-stage detection model containing a Backbone Network, a Region Proposal Network (RPN), and a series of Cascade Heads, as shown in Figure~\ref{fig:overall_architecture_cascade_rcnn}. The RPN is the first regression step of a two-stage detection model responsible for generating region proposals. More specifically, a set of predefined anchor boxes are defined and slides across the feature map to generate the fix-length of feature vectors for the classification and regression tasks in the RPN~\cite{ren2015faster}. The classification task classifies anchor boxes into object and background, and the regression task coarsely regresses the anchor boxes to generate higher-quality region proposals. Since the RPN only coarsely classifies and regresses the anchor boxes, the parameters of defining anchor boxes play a key role in the performance of the RPN, such as the number of anchor boxes, the aspect ratios of anchor boxes, and the scales of applied feature maps.

The Backbone Network is used to extract features of the input images, which is often followed by Feature Pyramid Network (FPN)~\cite{lin2017feature} to extract and fuse features from different scales. The extracted features, together with the region proposals generated by the RPN, are fed into the first cascade head for the classification and regression tasks,  and the regression results would be the inputs of the subsequent Cascade Head, as shown in Figure~\ref{fig:overall_architecture_cascade_rcnn}. 

Since there are multiple Cascade Heads, all the outputs of these Cascade Heads are used to calculate the loss. Moreover, the final loss of the model can be defined as the sum of these Cascade Heads loss and the RPN loss, as defined by Equation~\ref{eq:total_loss}, where $N$ is the number of Cascade Heads. It is worth mentioning that we follow the most popular Cascade R-CNN model to show three Cascade Heads in Figure~\ref{fig:overall_architecture_cascade_rcnn}. Each Cascade Head has a REG Head and a CLS Head for the regression and classification tasks, respectively. The input features of these REG Heads and CLS Heads $e_{cls}, e_{reg}$ are extracted by applying ROI Pooling operations to the features from Backbone Network with the proposal boxes $b$, which can be defined by Equations~\ref{eq:cascade_rcnn_cls} and ~\ref{eq:cascade_rcnn_reg} where $PROJ, ROI\_POOL$, and $BACKBONE$ are the Projection layer, ROI Pooling operations, and the Backbone Network. Therefore, for a trained model, the input features of the CLS Heads $e_{cls}$ are determined by the input image $x$ and the proposal boxes $b$, meaning that a single proposal box cannot be classified into multiple classes because CLS Heads are not multi-label classifiers. 

\begin{equation}
\label{eq:total_loss}
\mathcal{L} = \mathcal{L}_{rpn} +  \sum_{i=1}^{N} (\mathcal{L}^i_{cls} + \mathcal{L}^i_{reg})
\end{equation}


\begin{equation}
\label{eq:cascade_rcnn_cls}
e_{cls} = PROJ_{cls}(ROI\_POOL(BACKBONE(x), b))
\end{equation}

\begin{equation}
\label{eq:cascade_rcnn_reg}
e_{reg} = PROJ_{reg}(ROI\_POOL(BACKBONE(x), b))
\end{equation}

\subsubsection{Sparse R-CNN}
\label{sec:sparse_r_cnn}

\begin{figure*}[htp]
\begin{center}
\includegraphics[width=\textwidth, keepaspectratio]{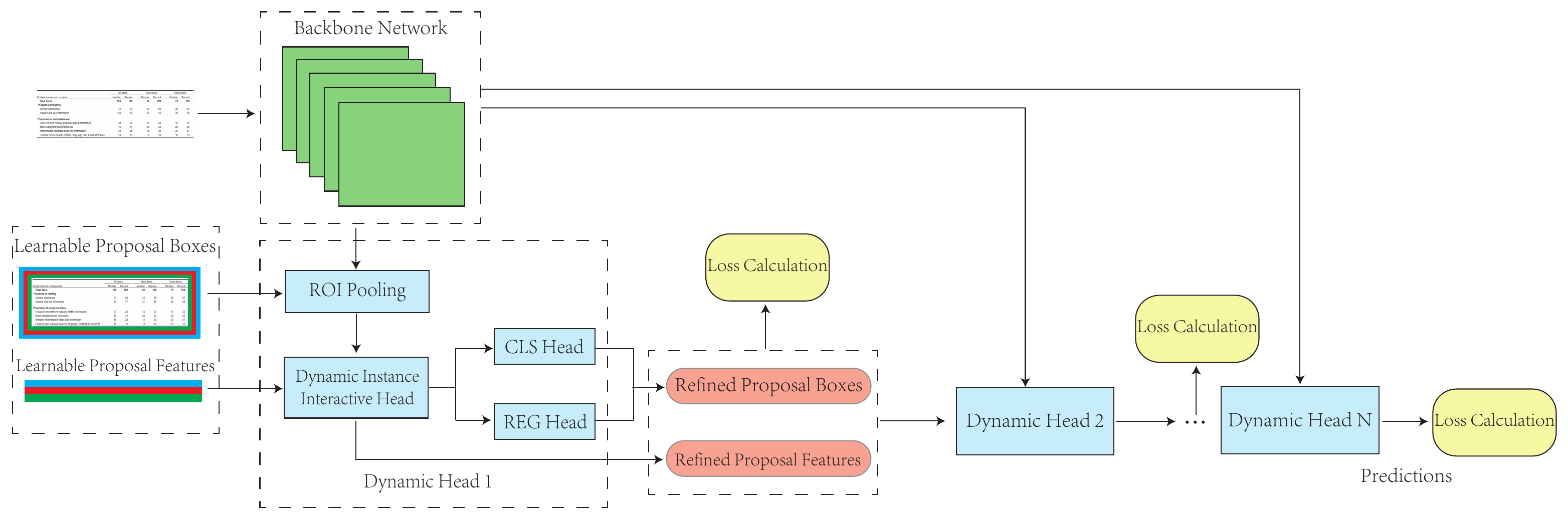}
\caption{Overall architecture of Sparse R-CNN.}
\label{fig:overall_architecture_sparse_rcnn}
\end{center}
\end{figure*}

Sparse R-CNN is a popular end-to-end transformer-based detection model. Similar to Cascade R-CNN, Sparse R-CNN also employs a cascade architecture containing a series of Dynamic Heads, as shown in Figure~\ref{fig:overall_architecture_sparse_rcnn}. In each Dynamic Head, an ROI Pooling layer is applied to extract features from the feature map based on the given proposal boxes, and the extracted features, together with the learnable proposal features, are fed to the Dynamic Instance Interactive Head to generate final features for the classification and regression tasks. Therefore, the features fed into CLS Head and REG Head of each Dynamic Head can be defined as Equations~\ref{eq:sparse_rcnn_cls} and ~\ref{eq:sparse_rcnn_reg}, where $BACKBONE$, $DYN\_HEAD$, and $PROJ$ are the Backbone Network, Dynamic Instance Interactive Head and the Projection layer, respectively, and $x$, $b$, $f$ are the input image, the proposal boxes and the learnable proposal features. It is worth mentioning that Sparse R-CNN does not use any RPN network to generate regional proposals. Instead, it proposes to use a set of learnable proposal boxes paired with a set of learnable features, in which learnable proposal boxes can be initialized by some pre-defined methods, such as image size initialization, random initialization, and grid initialization. Once the model is trained, the proposal boxes can be treated as an identical value, such as the box of image size, and their classification and regression results are mainly determined by their corresponding learnable proposal features $f$ and the input image $x$. Therefore, for a multi-label detection problem, when objects belonging to different classes can share an identical box, the learnable proposal features can be different for these objects, making it possible for Sparse R-CNN to deal with multi-label detection tasks.

\begin{equation}
\label{eq:sparse_rcnn_cls}
e_{cls} = PROJ_{cls}(DYN\_HEAD (ROI\_POOL(BACKBONE(x), b), f))
\end{equation}

\begin{equation}
\label{eq:sparse_rcnn_reg}
e_{reg} = PROJ_{reg}(DYN\_HEAD (ROI\_POOL(BACKBONE(x), b), f))
\end{equation}

\subsection{Rethinking Problem Formulations}
\label{sec:revisiting_problem_formulations}

\begin{figure}
     \centering
     \begin{subfigure}[b]{\columnwidth}
         \centering
         \includegraphics[width=0.9\columnwidth]{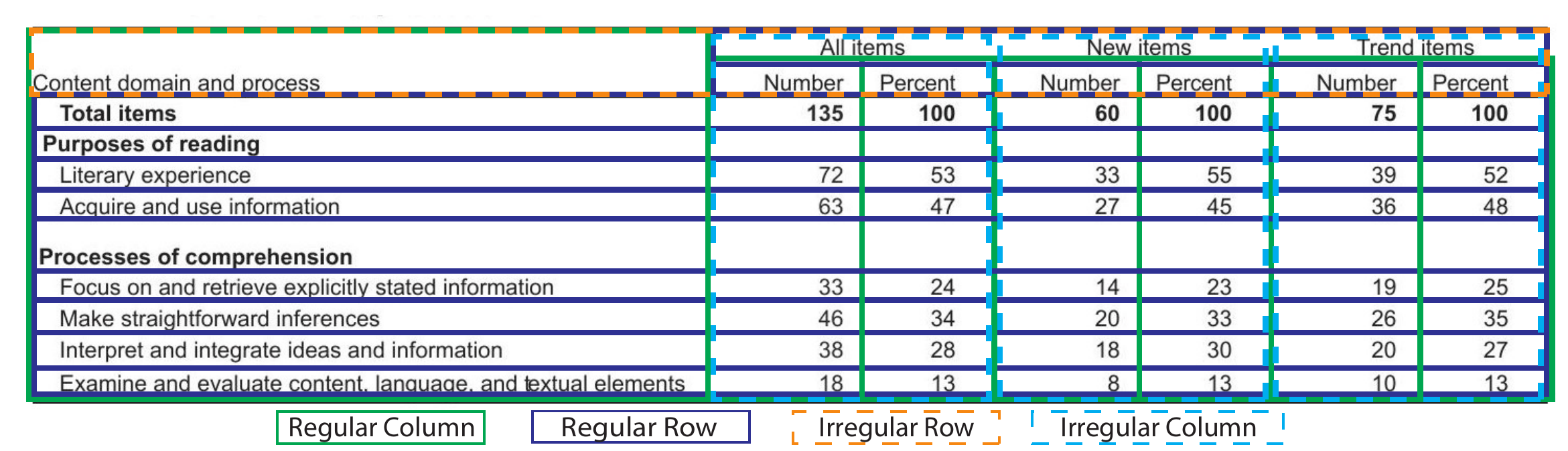}
         \caption{Four types of defined table components in TableStrRec~\cite{fernandes2023tablestrrec}. This definition can infer Spanning Cells but cannot provide Column Header information. Besides, the Projected Row Headers are treated as regular Rows, which can lead to a wrong structure.}
         \label{fig:definition_2classes}
     \end{subfigure}
     \hfill
     \begin{subfigure}[b]{\columnwidth}
         \centering
         \includegraphics[width=0.9\columnwidth]{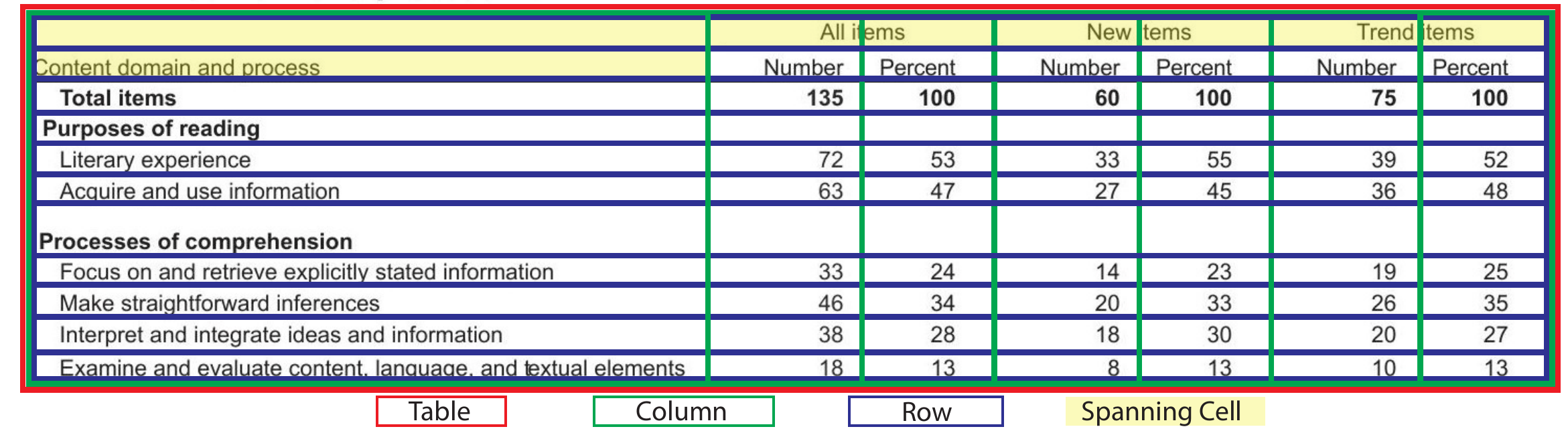}
         \caption{Four types of defined table components in the study~\cite{xiao2022efficient}. This definition cannot provide Column Header information, and the Projected Row Headers are treated as regular Rows, which can lead to a wrong structure.}
         \label{fig:definition_4classes}
     \end{subfigure}     
     \hfill
     \begin{subfigure}[b]{\columnwidth}
         \centering
         \includegraphics[width=0.9\columnwidth]{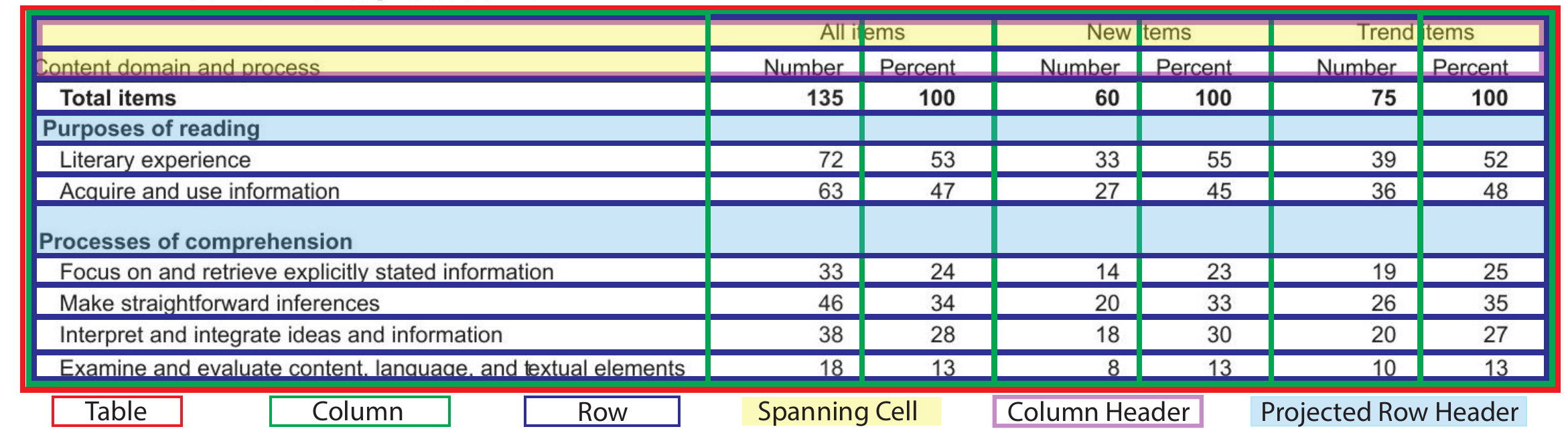}
         \caption{Six types of defined table components in PubTables1M~\cite{smock2022pubtables}. The defined Projected Row Headers in this sample share identical bounding boxes with two corresponding rows.}
         \label{fig:definition_6classes}
     \end{subfigure}
     \hfill
     \begin{subfigure}[b]{\columnwidth}
         \centering
         \includegraphics[width=0.5\columnwidth]{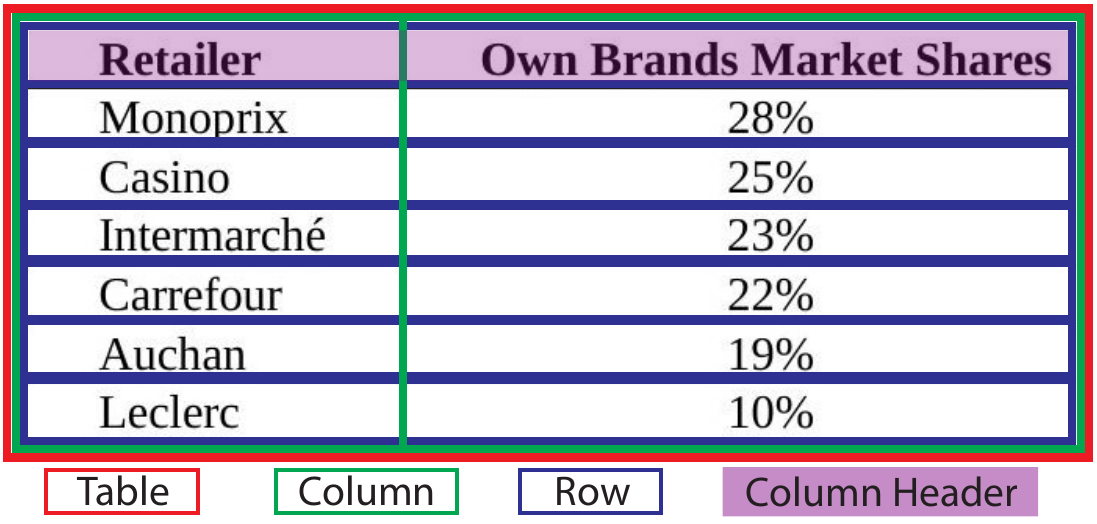}
         \caption{Six types of defined table components in PubTables1M~\cite{smock2022pubtables}. In this sample, the defined Column Header shares an identical bounding box with a defined Row. It is worth mentioning that there are no Spanning Cell and Projected Row Header in this sample.}
         \label{fig:definition_6classes_2}
     \end{subfigure}
     
    \caption{Different problem formulations for the detection-base TSR.}
    \label{fig:problem_defintion_comparison}
\end{figure}

As aforementioned in section~\ref{sec:introduction}, there have been many detection-based solutions~\cite{siddiqui2019deeptabstr,hashmi2021guided,fernandes2023tablestrrec,xiao2022efficient,smock2022pubtables} with different problem formulations that either oversimplified the TSR task or ignored its multi-label characteristic. More specifically, following image-to-sequence TSR models, a detection-based TSR model should be able to fully reconstruct the structure of both regular and spanning table cells, as well as provide information regarding header cells. However, studies~\cite{siddiqui2019deeptabstr, hashmi2021guided} formulate the problem as only detecting columns and rows, making them impossible to deal with spanning cells and identify header cells. TableStrRec~\cite{fernandes2023tablestrrec} further extend the formulation by defining regular column, regular row, irregular column, and irregular row so that the spanning cell can be inferred from these four types of components, as shown in Figure~\ref{fig:definition_2classes}. But still, this formulation cannot provide information regarding header cells, which is still over-simplied the TSR task. Study~\cite{xiao2022efficient} simplifies the formulation of in PubTables1M~\cite{smock2022pubtables}, directly detecting table, column, row, and spanning cell, as shown in Figure~\ref{fig:definition_4classes}, which is another formulation ignoring header cells' information. Besides, these three formulations treat the Projected Row Header as a regular row, resulting in over-simplified table structures. By contrast, PubTables1M~\cite{smock2022pubtables} defines six types of components, including Table, Column, Row, Spanning Cell, Column Header, and Projected Row Header, as shown in Figures~\ref{fig:definition_6classes} and ~\ref{fig:definition_6classes_2}, which can provide as much structure information as image-to-sequence TSR models. However, this formulation does not consider that some Column Headers and Projected Row Headers can share identical bounding boxes with corresponding Rows. For example, as shown in Figure~\ref{fig:definition_6classes}, the bounding boxes of the two Projected Row Headers can also be classified as Rows. Similarly, as shown in Figure~\ref{fig:definition_6classes_2}, the Column Header's bounding box is also the Row's bounding box. Therefore, the problem definition of study~\cite{smock2022pubtables} is a multi-label detection problem, which is must be considered when we choose and design detection models. It is worth mentioning that all these problem formulations use extracted table images as inputs. Even though many studies~\cite{cascadetabnet2020, xiao2023revisiting,xiao2023table} have achieved very promising performance on the Table Detection (TD) task, it is still difficult to guarantee that all the table content can be fully included in the detection results. Therefore, in practice, the detected bounding boxes of tables from the TD model are often padded with extra pixels, making it necessary to define a Table component for TSR.


\subsection{Revisiting Region Proposal Generation}
\label{sec:revisiting_region_proposal_generation}
\begin{figure}
     \begin{subfigure}[b]{0.5\columnwidth}
         \centering
         \includegraphics[width=\columnwidth]{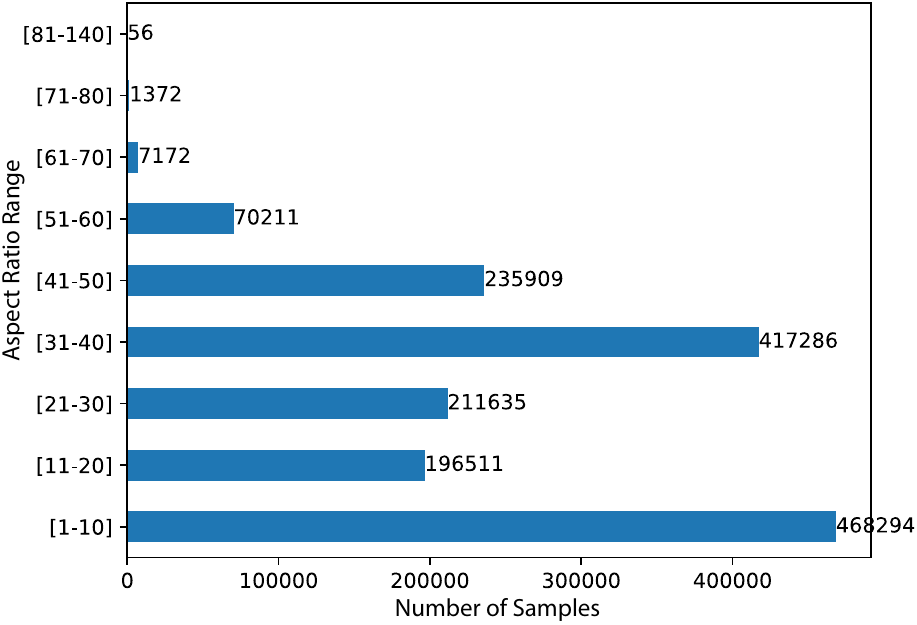}
         \caption{Aspect ratios of the FinTabNet training set.}
         \label{fig:aspect_ratio_fintabnet}
     \end{subfigure}
     \hfill
     \begin{subfigure}[b]{0.5\columnwidth}
         \centering
         \includegraphics[width=\columnwidth]{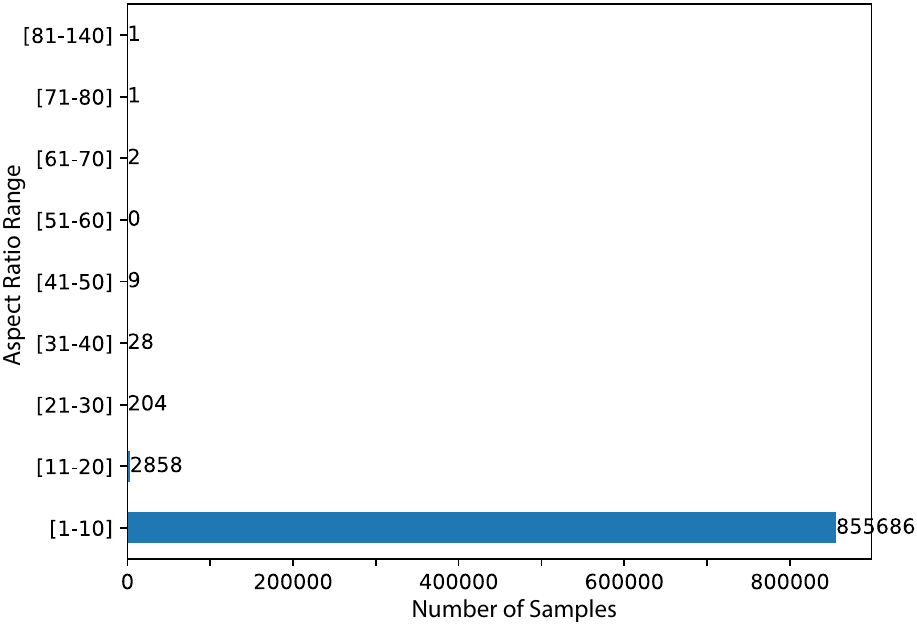}
         \caption{Aspect ratios of the COCO training set.}
         \label{fig:aspect_ratios_coco}
     \end{subfigure}
    \caption{Statistics of aspect ratio values of COCO and FinTabNet training sets. When an aspect ratio is less than 1, its multiplicative inverse counts the number of aspect ratios.}
    \label{fig:aspect_ratios_coco_fintabnet}
\end{figure}

As aforementioned in section~\ref{sec:prelimiaries}, the parameters of generating anchor boxes in the RPN play a key role in two-stage detection models, while transformer-based detection models, such as DETR and Sparse R-CNN, use learnable queries or proposals without the need to tune the RPN. If we choose two-stage detection models, such as Cascade R-CNN which is used in TableStrRec~\cite{fernandes2023tablestrrec}, we need to identify the difference between the TSR detection problem and widely discussed common object detection problem, because the default settings of detection frameworks, such as Detectron2~\cite{wu2019detectron2} and MMDetection~\cite{mmdetection} are often tuned on COCO~\cite{lin2014microsoft} dataset. Therefore, we compare the statistics of the COCO dataset with a popular TSR dataset, FinTabNet~\cite{zheng2020global}, regarding the number of objects in each image and the aspect ratios of objects. More specifically, the COCO training set contains 118287 images and 860001 target objects, resulting in an average of 7.27 objects in each image, while the FinTabNet training set contains 78537 images, 1628298 target objects, resulting in an average of 20.73 objects in each image. Besides, the aspect ratios of objects in these two dataset are also very different, as shown in Figure~\ref{fig:aspect_ratios_coco_fintabnet}. The vast majority of target objects in the COCO training set have aspect ratios between 1 and 10, while objects in the FinTabNet training set have much larger aspect ratios. Therefore, we need to consider these differences when tuning the parameters of RPN if we apply a two-stage object detection model for the TSR task, such as increasing the number of region proposals and adjusting the aspect ratios of anchor boxes. On the other hand, transformer-based detection models, such as Sparse R-CNN and DETR, can alleviate the issues caused by these differences intrinsically because they use learnable queries (learnable proposals) instead of an RPN, as discussed in section~\ref{sec:prelimiaries}. However, increasing the number of learnable queries for each image might be also useful for transformer-based detection models because TSR datasets contain more objects than common object detection datasets.

\subsection{Rethinking Detection and TSR Metrics}
\label{sec:rethinking_detection_metrics_tsr_metrics}
As mentioned in sections~\ref{sec:introduction} and ~\ref{sec:related_work}, detection-based TSR models need a deterministic rule-based post-processing method to transform the detected objects into structured sequences. Existing studies~\cite{siddiqui2019deeptabstr, hashmi2021guided, fernandes2023tablestrrec, xiao2022efficient} usually use the detection performance to evaluate the model performance before applying the post-processing method. However, the detection metrics are not aligned with cell-level TSR metrics. We use COCO~\cite{lin2014microsoft} and TEDS~\cite{zhong2020image} metrics as examples for further analysis in this section. The COCO metrics employ mean Average Precision (mAP) to evaluate the model performance, which can be defined by Equation~\ref{eq:m_ap} where $N$, $precision_i(r)$ and $dr$ in Equation~\ref{eq:m_ap} are the number of classes, and the precision at a given recall level $r$ for class $i$. In practice, the precision-recall curves in COCO metrics are computed for each class at a series of IoU thresholds, and the integral of $precision_i(r)$ often is approximated by the discrete sum. The IoU score can be defined by Equation ~\ref{eq:iou}, where $A \cap B$, $A \cup B$ are the intersection and union of bounding boxes $A$ and $B$. In practice, in many studies, mAP is represented by AP and calculated by averaging the mean precision scores of all categories at IoU thresholds from 0.5 to 0.95 with 0.05 intervals. AP50, AP75 are the mean precision scores of all categorizes at IoU thresholds 0.5 and 0.75, respectively. Therefore, COCO metrics are IoU-based evaluation metrics. By contrast, TEDS can be defined by Equation~\ref{eq:teds_score}, where EditDist is the tree-edit distance, and $T$ is the number of nodes in the tree, meaning that TEDS is not correlated with IoU scores.

\begin{equation}
    \label{eq:m_ap}
    mAP = \frac{1}{N} \sum_{i=1}^{N} \left( \int_{0}^{1} \text{precision}_i(r) \, dr \right)
\end{equation}

\begin{equation}
    \label{eq:iou}
    IoU = \frac{A \cap B}{A \cup B}
\end{equation}

\begin{equation}
\label{eq:teds_score}
TEDS(T_a, T_b) = 1 - \frac{EditDist(T_a, T_b)}{max(\mid T_a \mid, \mid T_b \mid)}
\end{equation}

\begin{figure*}[htp]
\begin{center}
\includegraphics[width=\textwidth, keepaspectratio]{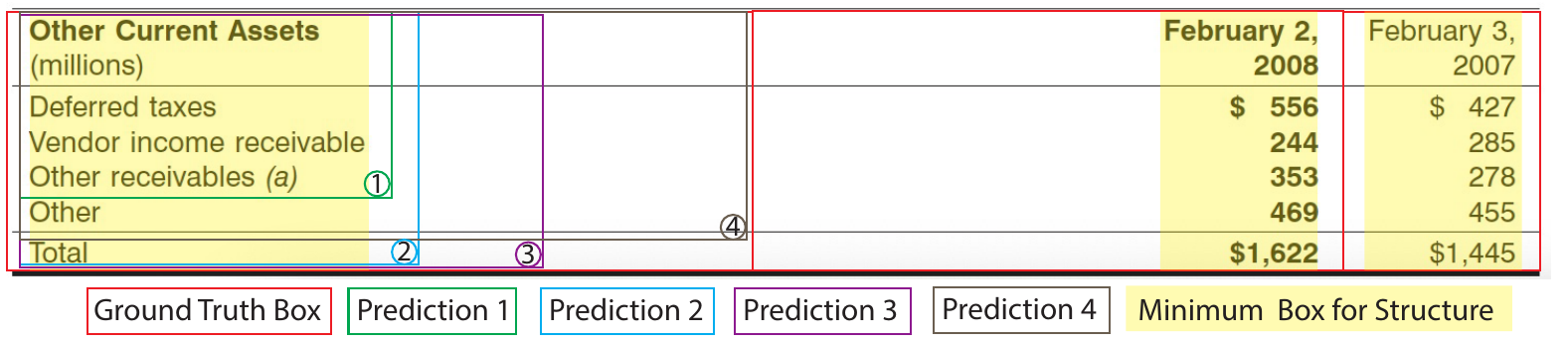}
\caption{A sample from the FinTabNet dataset with ground truth boxes larger than the minimum bounding boxes for table structure. We only show the annotations of Columns for simplicity.}
\label{fig:metrics_misalignment_sample}
\end{center}
\end{figure*}

On the other hand, TSR datasets usually use a canonicalization procedure~\cite{smock2022pubtables} or annotate the bounding boxes following the lines in tables, which makes the ground truth boxes larger than the minimum box that can recover the structure of the table. Figure~\ref{fig:metrics_misalignment_sample} shows an example from the FinTabNet dataset, whose ground truth boxes are larger than the minimum bounding boxes for table structure. Considering the four prediction boxes in Figure~\ref{fig:metrics_misalignment_sample}, since the prediction 1 is smaller than the minimum box for table structure, and the prediction2 can cover all content of the minimum box for table structure and has a larger IoU with the ground truth box, prediction 2 can lead to better performance regarding both COCO and TEDS metrics than prediction 1. By contrast, prediction 3 has a larger IoU with the ground truth box than prediction 2, which can lead to better detection performance. However, when it comes to TEDS, prediction 3 cannot show any superiority compared to prediction 2, because both of them can cover the minimum box for table structure. When we compare prediction 2 and prediction 4, prediction 4 has a larger IoU with the ground truth box, making it better on detection performance, but it loses information of the row, making its performance in TEDS worse than prediction 2. Therefore, because of the definitions of COCO and TEDS metrics and the procedure of annotating datasets, a detection-based TSR model might be over-optimized towards detection performance without increasing the TEDS performance and sometimes can decrease the TEDS performance.

\subsection{Rethinking Feature Extraction}
\label{sec:rethinking_feature_generation}
\begin{figure*}[htp]
\begin{center}
\includegraphics[width=\textwidth, keepaspectratio]{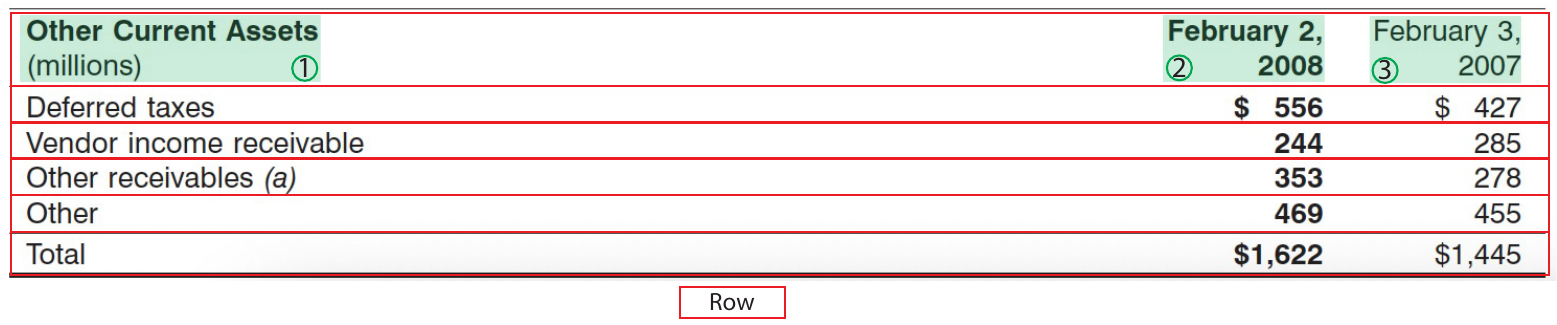}
\caption{A sample from the FinTabNet dataset. We only show its Row annotations for simplicity.  The first Row in this Figure contains three major parts numbered 1 to 3.}
\label{fig:long_dependency_sample}
\end{center}
\end{figure*}

As mentioned in section~\ref{sec:introduction}, deformable convolution~\cite{dai2017deformable} has been applied in detection-based TSR~\cite{fernandes2023tablestrrec, siddiqui2019deeptabstr} and other related solutions~\cite{siddiqui2018decnt,mondal2023dataset}, demonstrating its effectiveness in improving detection performance. Deformable convolution uses a learnable grid offset to sample the grid points from the feature map and then apply the convolution operation to the sampled grid points, as defined by Equation~\ref{eq:deformable_conv},

\begin{equation}
\label{eq:deformable_conv}
  \mathbf{z}_{p_0}= \sum_{p_n \in R} {w(p_n)x(p_0 + p_n + \Delta p_n) }     
\end{equation}
where $p_0$ is the location on the output feature map $\mathbf{z}$, $p_n$ is the $n$th grid point in grid $R$, and $\Delta p_n$ is the $n$th learnable offset. Since the offset $\Delta p_n$ applied to the deformable convolution is usually obtained by a regular convolution with small kernels, such as a 3 * 3 kernel, it can only improve the local feature instead of building long-range dependencies. However, building the long-range dependencies for the TSR task is important because of the characteristics of table components. More specifically, different parts of a single table component are often sparsely distributed across the table instead of a single area of compact pixels like common objects. Figure~\ref{fig:long_dependency_sample} shows a sample with its Row annotations. Taking the first Row as an example, as shown in Figure~\ref{fig:long_dependency_sample}, it mainly contains three parts, which are distributed sparsely, and there is a large space between the first part and the second part, even they all belong to a single target component. Therefore, it is important to build long-range dependencies together with improving local features, such as applying deformable convolution. And over-optimized local features, such as merely applying deformable convolution might degrade the performance regarding the TEDS.

\section{Proposed Method}
\label{sec:proposed_method}
In this section, we demonstrate how to fill the performance gap between detection-based and other types of TSR models by applying very simple methods to tailor the Cascade R-CNN model based on our analysis and findings in the previous sections. We first introduce the our problem formulation, then give the details of the proposed methods, including adjusting the parameters of the RPN, applying deformable convolution and introducing Spatial Attention Module.  

\subsection{Proposed Problem Formulation}
\label{sec:pseudo_class_generation}
\begin{figure}
     \centering
     \begin{subfigure}[b]{\columnwidth}
         \centering
         \includegraphics[width=0.9\columnwidth]{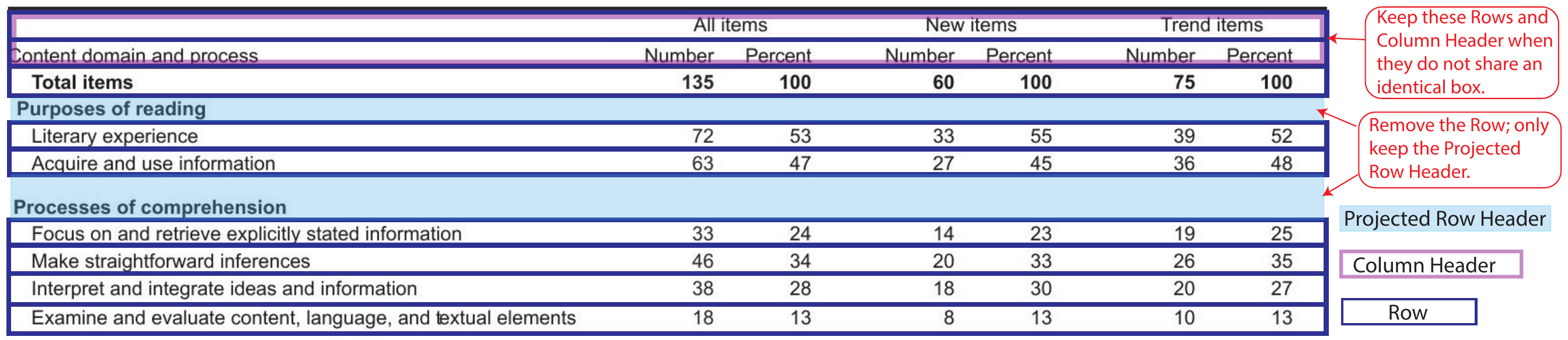}
         \caption{An example of our problem formulation. In this example, we remove two Rows because their bounding boxes are identical with two Projected Row Headers.}
         \label{fig:new_definition}
     \end{subfigure}
     \hfill
     \begin{subfigure}[b]{\columnwidth}
         \centering
         \includegraphics[width=0.7\columnwidth]{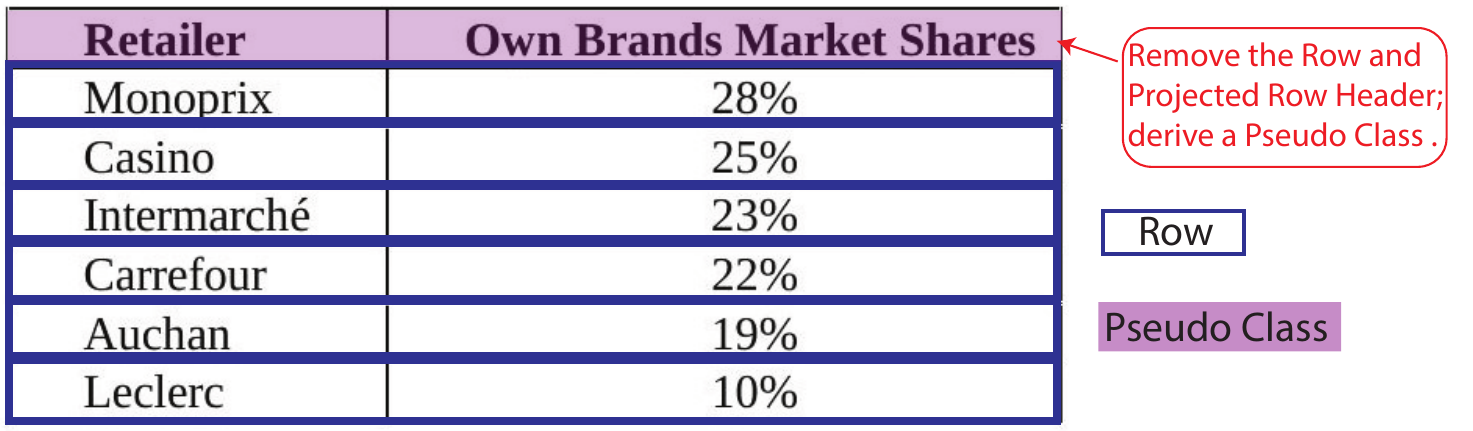}
         \caption{An example of our problem formulation. In this example, we derive a Pseudo Class because its bounding box simultaneously belongs to a Row and a Column Header.}
         \label{fig:new_definition_2}
     \end{subfigure}
     
    \caption{Examples of our proposed problem formulation. Since the definitions of Table, Column, and Spanning Cells are same with PubTables1M, only Row, Column Header and Projected Row Header are showed for simplicity.}
    \label{fig:new_formulation}
\end{figure}

As mentioned in sections~\ref{sec:prelimiaries} and ~\ref{sec:revisiting_problem_formulations}, the definition of PubTables1M~\cite{smock2022pubtables} can provide as much information as other types of solutions and is a multi-label detection problem, which is challenging for two-stage detectors. Therefore, we follow PubTables1M to define six table components: Table, Column, Row, Spanning Cell, Projected Row Header, and Column Header, and transform the formulation into a single-class detection problem. More specifically, we remove the Rows that share their bounding boxes with the Projected Row Header, as shown in Figure~\ref{fig:new_definition}, and use a Pseudo Class to replace the Rows and Column Headers when they share identical bounding boxes, as shown in Figure~\ref{fig:new_definition_2}. It is worth mentioning that only the Row, Projected Row Header, and Column Header are shown because the Table, Column, and Spanning Cell are the same as PubTables1M. These two samples are also in Figures~\ref{fig:definition_6classes} and ~\ref{fig:definition_6classes_2}, which show their original definition in PubTables1M. 

Formally, the ground truth $Y$ in PubTable1M's definition for each image is a set of tuples containing bounding boxes and their corresponding labels, as defined by Equation~\ref{eq:multi_class_definition}, where $b_i, c_i$ are the $i$th bounding box and its class, and values from $0$ to $5$ are the defined Table, Column, Row, Spanning Cell, Projected Row Header and Column Header, respectively. 

\begin{equation}
\label{eq:multi_class_definition}
Y = \{(b_i, c_i)\}_{i=1}^N, c_i \in \{0, 1, 2, 3, 4, 5\}, \forall i \neq j, (c_i \neq c_j) \land ((b_i = b_j) \lor (b_i \neq b_j))
\end{equation}

By contrast, in this study, considering the observation that the defined Projected Row Headers are all Rows at the same time, we only keep the Projected Row Headers samples during the training. Since some Column Headers can share identical bounding boxes with corresponding Rows, we derive a pseudo class for these overlapped samples and remove the original overlapped samples. Therefore, during the training stage, we refactor the ground truth for each image to the regular single-label classification, as defined by Equation~\ref{eq:single_class_definition}, where values $0$ to $6$ are the Table, Column, Row, Spanning Cell, Projected Row Header, Column Header and the Pseudo Class, respectively. During the testing stage, the results of Project Column Header are duplicated once to generate their corresponding prediction Rows, and the results of the pseudo-class are duplicated twice to generate the corresponding prediction Rows and Headers, so that we can still follow the formulation defined by Equation~\ref{eq:multi_class_definition} to evaluate the model performance.

\begin{equation}
\label{eq:single_class_definition}
Y = \{(b_i, c_i)\}_{i=1}^N, c_i \in \{0, 1, 2, 3, 4, 5, 6\}, \forall i \neq j, (c_i \neq c_j) \land (b_i \neq b_j)
\end{equation}

\subsection{Tuning Parameters of RPN}
\label{sec:aspect_ratio_tuning}
As mentioned in sections~\ref{sec:prelimiaries} and~\ref{sec:revisiting_region_proposal_generation}, regional proposal generation is a critical step in two-stage detectors, which need to be carefully considered for the TSR problem. Therefore, we adjust Aspect Ratios and increase the number of generated regional proposals for our tailored model. More specifically, aspect ratios control the shape of the generated anchor boxes. Popular implementations of Cascade R-CNN, such as Detectron2~\cite{wu2019detectron2}, usually use 0.5, 1.0, and 2.0 as default values, which can work well for detecting common objects, such as the objects in COCO~\cite{lin2014microsoft} dataset. However, in the context of TSR, the range of aspect ratios is much larger because of the shape of the table components, as discussed in section~\ref{sec:revisiting_region_proposal_generation}. Without proposing fancy new modules to select suitable values, we simply select the values based on the statistics of the training sets. Taking the FinTabNet dataset as an example, the aspect ratios of the defined components are shown in Figure~\ref{fig:aspect_ratios_coco_fintabnet}. The maximum value is 140, far larger than the popular choices in common object detection. Besides, the majority of aspect ratios in Figure~\ref{fig:aspect_ratios_coco_fintabnet} are in the range between 1 and 60. Therefore, we extend this parameter for our proposed model as [0.0125, 0.025, 0.0625, 0.125, 0.25, 0.5, 1.0, 2.0, 4.0, 8.0, 16, 40, 80]. A further detailed parameter table is provided in section~\ref{sec:experiments}. It is worth mentioning that when an aspect ratio is less than 1, its multiplicative inverse is applied to count the number of aspect ratios in Figure~\ref{fig:aspect_ratios_coco_fintabnet}. We also didn't further fine-tune this parameter through validation, which means that they might not be optimal. But this parameter has improved the model performance by around 2.7\% as shown in section~\ref{sec:ablation_study}. Besides, since increasing the number of proposals has been applied in existing studies~\cite{fernandes2023tablestrrec}, and demonstrated its effectiveness, we increase it for both the base Cascade R-CNN and our proposed model.

\subsection{Spatial Attention and Deformable Convolution}
\label{sec:spatial_attention_and_deformable}
\begin{figure*}[htp]
\begin{center}
\includegraphics[width=0.8\textwidth, keepaspectratio]{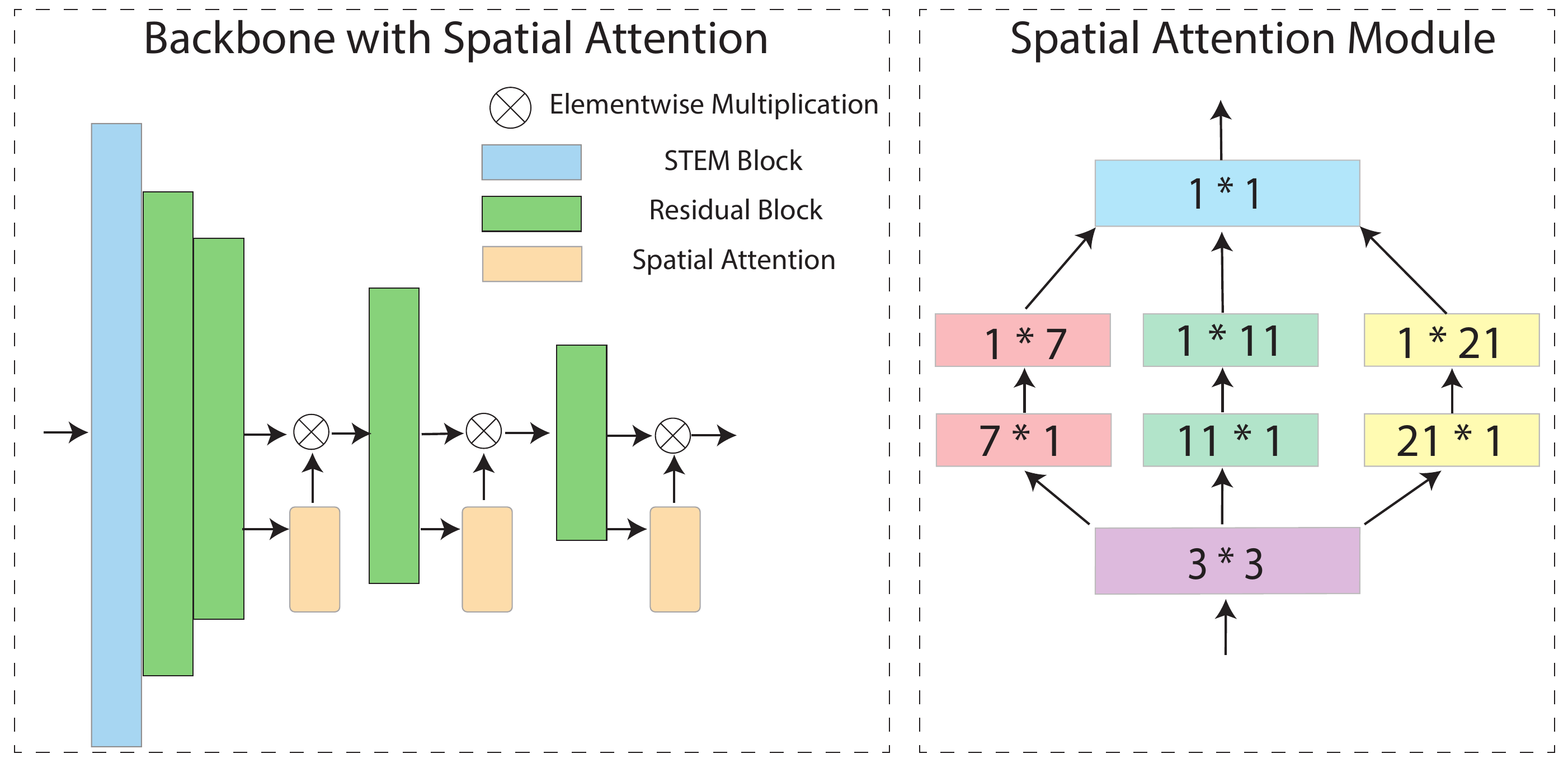}
\caption{Architecture of proposed Spatial Attention Module. A ResNet backbone consists of a STEM Block and four stages of Residual Block. Our proposed Spatial Attention Module are inserted between the blocks of the backbone to build long dependencies.}
\label{fig:spatial_attention}
\end{center}
\end{figure*}

As discussed in section~\ref{sec:rethinking_feature_generation}, building long-range dependencies for detecting the defined components is important. Inspired by the recent studies using large convolution kernels~\cite{ding2022scaling, liu2022convnet}, we introduce a Spatial Attention Module for our solution, whose architecture is shown in Figure~\ref{fig:spatial_attention}. For the design of the Spatial Attention Module, we use a similar architecture with MSCA~\cite{guo2022segnext} containing multiple branches and large kernel convolutions and use spatial and depthwise separable convolution~\cite{chollet2017xception,howard2017mobilenets} to reduce the number of parameters. More specifically, for the spatial separable convolution, we use a pair of $7 * 1$ and $1 * 7$ kernels to replace a typical $7 * 7$, use the pair of $11 * 1$ and $1 * 11$, and the pair of $21 \times 1$ and $1 \times 21$ to replace $11 * 11$ and $21 * 21$ kernels, respectively. For the depthwise separable convolution, we applied the convolution on each channel of the feature maps independently. Then, the outputs of the three branches are concatenated together as the input of a convolution layer with $1 \times 1$ kernel to make the channel dimension the same as the inputs. The proposed Spatial Attention Module can be easily inserted into the Backbone Network between two blocks because they do not change the feature shapes. For example, for a typical backbone network implemented by ResNet~\cite{he2016deep} containing a STEM block and four Residual Blocks, as shown in Figure~\ref{fig:spatial_attention}, the Spatial Attention Module can be inserted after the last three Residual Blocks to generate the spatial attention, then the spatial attention can be applied to the original outputs of each Residual Block by Element-wise Multiplication. It is worth mentioning that the Spatial Attention Module shown in Figure~\ref{fig:spatial_attention} have independent trainable parameters, and all the feature maps are padded correspondingly to keep the size of the feature maps. 

On the other hand, as discussed in section~\ref{sec:rethinking_feature_generation}, deformable convolution~\cite{dai2017deformable} can improve the local feature generation and has been demonstrated to help improve the detection performance on the document image detection tasks by many studies~\cite{fernandes2023tablestrrec,siddiqui2018decnt,mondal2023dataset}. Therefore, in this study, we apply the proposed Spatial Attention Module and deformable convolution to build long-range dependencies and improve local features together. 

\section{Experiments and Analysis}
\label{sec:experiments}

\subsection{Datasets and Experimental Settings} 
\label{sec:datasets}

We utilize four datasets in this work, including SciTSR~\cite{chi2019complicated},  FinTabNet~\cite{zheng2020global}, PubTabNet~\cite{zhong2020image} and PubTables1M~\cite{smock2022pubtables}. As discussed in the study~\cite{smock2023aligning}, FinTabNet and SciTSR datasets contain noise annotations that harm the model performance. Therefore, we use their cleaned versions proposed in the study in~\cite{smock2023aligning}. Each image sample in these four datasets contains only a table with extra padding pixels to ensure the entire table is extracted. The SicTSR dataset is collected from academic publications containing 7453, 1034, and 2134 samples for training, validation, and testing. PubTables1M dataset is a large-scale dataset for the TSR problem collected from the PMCOA corpus, containing 758849 training samples, 94959 validation samples, and 93834 testing samples. Since the PubTabNet dataset does not provide original PDF files, we cannot process it to make detection annotations. Besides, its testing is not publicly available. Therefore, we use its validation set to evaluate the model trained with the PubTable1M dataset. Following the study in~\cite{smock2023aligning}, we use the code base in~\cite{smock2021tabletransformer} to process the datasets and align the formats of these datasets. FinTabNet is also a large dataset widely used for the TSR problem, containing 78537, 9289, and 9650 samples for training, testing, and validation. FinTabNet is collected from the annual reports of companies, making its data source different from the other datasets. Table~\ref{tab:datasets_statis} summarizes the datasets used in this study for the model evaluation.

\begin{table*}[!htpb]
\centering
\caption{\label{tab:datasets_statis}Summary of datasets.} 
\begin{tabular}{ c c c c} 
\hline
Dataset & Train & Validation & Test \\
\hline
SciTSR~\cite{chi2019complicated} & 7,453 & 1,034 & 2,134 \\
FinTabNet~\cite{zheng2020global} & 78,537 & 9,650 & 9,289  \\
PubTabNet~\cite{zhong2020image} & 500,777 & 9,115 & - \\
PubTables1M~\cite{smock2022pubtables} & 758,849 & 94,959 & 93,834 \\
\hline
\end{tabular}
\end{table*}

Since the TSR problem in this study is formulated as an object detection problem, we use both detection and cell-level TSR metrics for the model evaluation. For the detection metric, we employ the widely accepted COCO metrics~\cite{lin2014microsoft}, which has been discussed in section~\ref{sec:rethinking_detection_metrics_tsr_metrics}. More specifically, mean Average Precision (AP), $AP50$, $AP75$, $AP_s$, $AP_m$, $AP_l$, and object-specific AP scores are used as metrics, where $AP50$, $AP75$ are the APs using 0.50 and 0.75 as IoU thresholds, respectively. $AP_s$, $AP_m$, and $AP_l$ are the APs of different target object sizes, defined by Equation~\ref{eq:object_scale}. 

\begin{equation}
\label{eq:object_scale}
  object\_size =
    \begin{cases}
      small & \text{if area \textless $ 32^2$ px}\\
      medium & \text{if $ 32^2$ \textless area \textless $ 64^2$ px} \\
      large &  \text{otherwise} \\
    \end{cases}       
\end{equation}

For the TSR metric, we choose structure-only Tree-Edit-Distance-Based Similarity(TEDS)~\cite{zhong2020image}, which is firstly introduced in the study to overcome the drawbacks of adjacency relation metrics~\cite{hurst2003constraint}, and can be defined as Equation~\ref{eq:teds_score} as discussed in section~\ref{sec:rethinking_detection_metrics_tsr_metrics}. We use structure-only TEDS in the study, which only considers the HTML tags without extracting their contents to avoid the influence of OCR tools. The testing samples can be categorized into simple and complex groups based on whether they have cells spanning multiple columns and rows.

\subsection{Implementation Details and Experimental Results}
\label{sec:expeirmental_results}
To verify the effectiveness of our proposed solution, we include all three types of state-of-the-art methods as discussed in section~\ref{sec:related_work}. For the detection-based methods, Cascade R-CNN~\cite{Cai2018CascadeRD}, Deformable-DETR~\cite{zhu2020deformable} and Sparse R-CNN~\cite{sun2021sparse} are used as benchmark models, in which Cascade R-CNN~\cite{Cai2018CascadeRD} is also the based model of the proposed methods, Deformable-DETR and Sparse R-CNN are two state-of-the-art transformer-based detection models. For the image-to-sequence models, EDD~\cite{zhong2020image}, TableFormer~\cite{nassar2022tableformer}, TableMaster~\cite{ye2021pingan}, VAST~\cite{huang2023improving} and MTL-TabNet~\cite{visapp23namly} are included. TSRFormer-DQ-DETR~\cite{WANG2023109817} and RobustTabNet~\cite{ma2023robust} are two state-of-the-art models following the pipeline of detecting separation lines and then merging cell grids, which can treated as a graph-based model as discussed in section~\ref{sec:related_work}. 
TSRNet~\cite{li2022table} is also a graph-based methods which detect table cells first, then applys GNN to build the relations among the detected cells. 

\begin{table}[ht!]
\caption{Key training parameters of the proposed model. MAX\_ITER and STEPS are for the FinTabNet dataset as examples.}
\centering
\begin{tabular}{ l | c | l  }
\hline
\label{table:modek_key_parameters} 
Parameter & Value & Description \\
\hline
RESNETS.NORM & nnSyncBN &  Batch Normalization for the Backbone Network  \\
MAX\_ITER & 112,500 &  total number of mini-batch  \\
STEPS & 84,375 & the mini-batch to apply the learning rate schedule  \\
SCHEDULER & MultiStepLR & the scheduler to change the learning rate \\
NMS\_THRESH & 0.9 & non-maximum suppression threshold\\
PRE\_NMS\_TOPK\_TRAIN & 4000 & RPN proposals to keep before applying NMS in training\\
PRE\_NMS\_TOPK\_TEST & 2000 & RPN proposals to keep before applying NMS in testing\\
POST\_NMS\_TOPK\_TRAIN & 4000 & RPN proposals to keep after applying NMS in training\\
POST\_NMS\_TOPK\_TEST & 2000 & RPN proposals to keep after applying NMS in testing\\
DEFORM\_ON\_PER\_STAGE & [True, True, True, True] & whether to use deformable convolution in each backbone stage\\
\hline
\end{tabular}
\end{table}

We implement Cascade R-CNN and our proposed method based on the Detection2~\cite{wu2019detectron2},  the Deformable-DETR based on detrex~\cite{ren2023detrex}, and the Sparse R-CNN with their official codebase. For the Deformable-DETR and Sparse R-CNN, we use their default parameters. For the Cascade R-CNN baseline, we align the number of regional proposals and the batch normalization method to the TSRDet, as shown in Table~\ref{table:modek_key_parameters}. All these detection models are using ResNet50~\cite{he2016deep} pre-trained with ImageNet~\cite{deng2009imagenet} as the backbone network. We also re-train TableMaster~\cite{ye2021pingan} with the FinTabNet dataset based on its official code base. We term our proposed method with TSRDet for fast reference. For the implementation of the proposed TSRDet, aspect ratios in the anchor box generation are set as [0.0125, 0.025, 0.0625, 0.125, 0.25, 0.5, 1.0, 2.0, 4.0, 8.0, 16, 40, 80], and other key parameters are summarized in Table~\ref{table:modek_key_parameters}. Notably, to calculate the structure-only TEDS, we use the scripts provided by study in~\cite{smock2021tabletransformer} to generate the HTML sequences from the detected components, and all benchmark models, except our proposed model, are using the original definition of PubTables1M, which treats all table components independently with its multi-label detection setting. All the models are trained with 240, 120, and 60 epochs for the SciTSR, FinTabNet, and PubTables1M datasets, respectively.

\begin{table}[!htpb]
\centering
\caption{Experimental results on SciTSR dataset with structure-only TEDS score. Sim. means the tables without spanning cells and Com. represents the tables with spanning cells.}
\begin{tabular}{ c c c c} 
\hline
\multirow{2}{*}{Model} & \multicolumn{3}{c}{TEDS-struc.(\%)} \\
& Sim. & Com. & All  \\
\hline
 Cascade R-CNN & 77.31 & 84.74 & 79.09  \\
 Deformable-DETR & 98.17 & 94.59 & 97.30 \\
 Sparse R-CNN & $\mathbf{99.08}$ & 95.92 & 98.30 \\
 TSRDet(Ours) & 98.59 & $\mathbf{97.88}$ & $\mathbf{98.41}$ \\
\hline
\end{tabular}
\label{tab:scitsr_teds}
\end{table}

\begin{table}[!htpb]
\centering
\caption{Experimental results on FinTabNet dataset with structure-only TEDS score. Sim. means the tables without spanning cells and Com. represents the tables with spanning cells.}
\begin{tabular}{ c c c c } 
\hline
\multirow{2}{*}{Model} & \multicolumn{3}{c}{TEDS-struc.(\%)} \\
& Sim. & Com. & All \\
\hline
 EDD~\cite{zhong2020image} & 88.40 & 92.08 & 90.60\\
 TableFormer~\cite{nassar2022tableformer} & 97.50 & 96.00 & 96.80 \\
 TableMaster~\cite{ye2021pingan} & 98.36 & 98.28 & 98.32 \\
 VAST~\cite{huang2023improving} & - & - & 98.63 \\
 MTL-TabNet~\cite{visapp23namly} & 99.07 & 98.46 & 98.79 \\
 TSRFormer-DQ-DETR~\cite{WANG2023109817} & - & - & 98.40 \\
\hline
 Cascade R-CNN & 82.17 & 92.50 & 87.49 \\
 Deformable-DETR & 98.08 & 97.54 & 97.81 \\
 Sparse R-CNN & 98.36 & 97.91 & 98.13 \\
 TSRDet(Ours) & $\mathbf{99.08}$ & $\mathbf{99.02}$ & $\mathbf{99.05}$ \\
\hline
\end{tabular}
\label{tab:fintabnet_teds}
\end{table}

\begin{figure}
     \centering
     \begin{subfigure}[b]{\columnwidth}
         \centering
         \includegraphics[width=0.6\columnwidth]{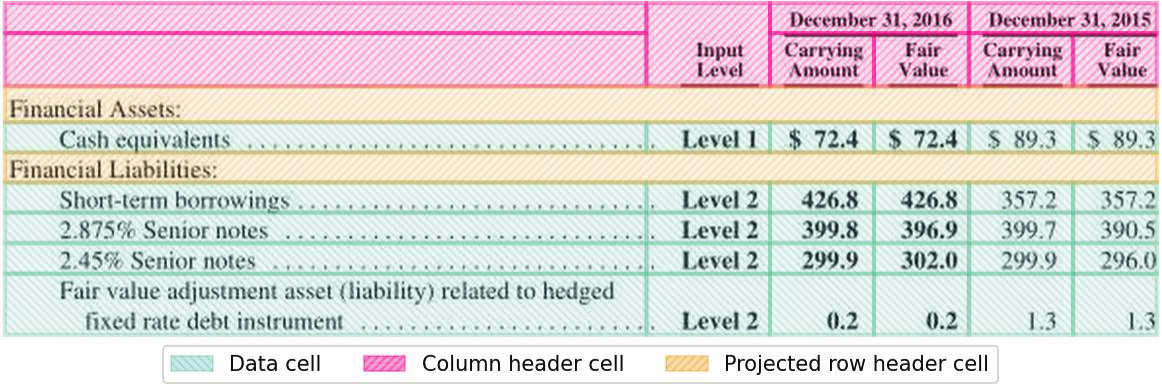}
         \caption{A prediction sample after post-processing.}
         \label{fig:tsr_result_vis}
     \end{subfigure}
     \hfill
     \begin{subfigure}[b]{\columnwidth}
         \centering
         \includegraphics[width=0.6\columnwidth]{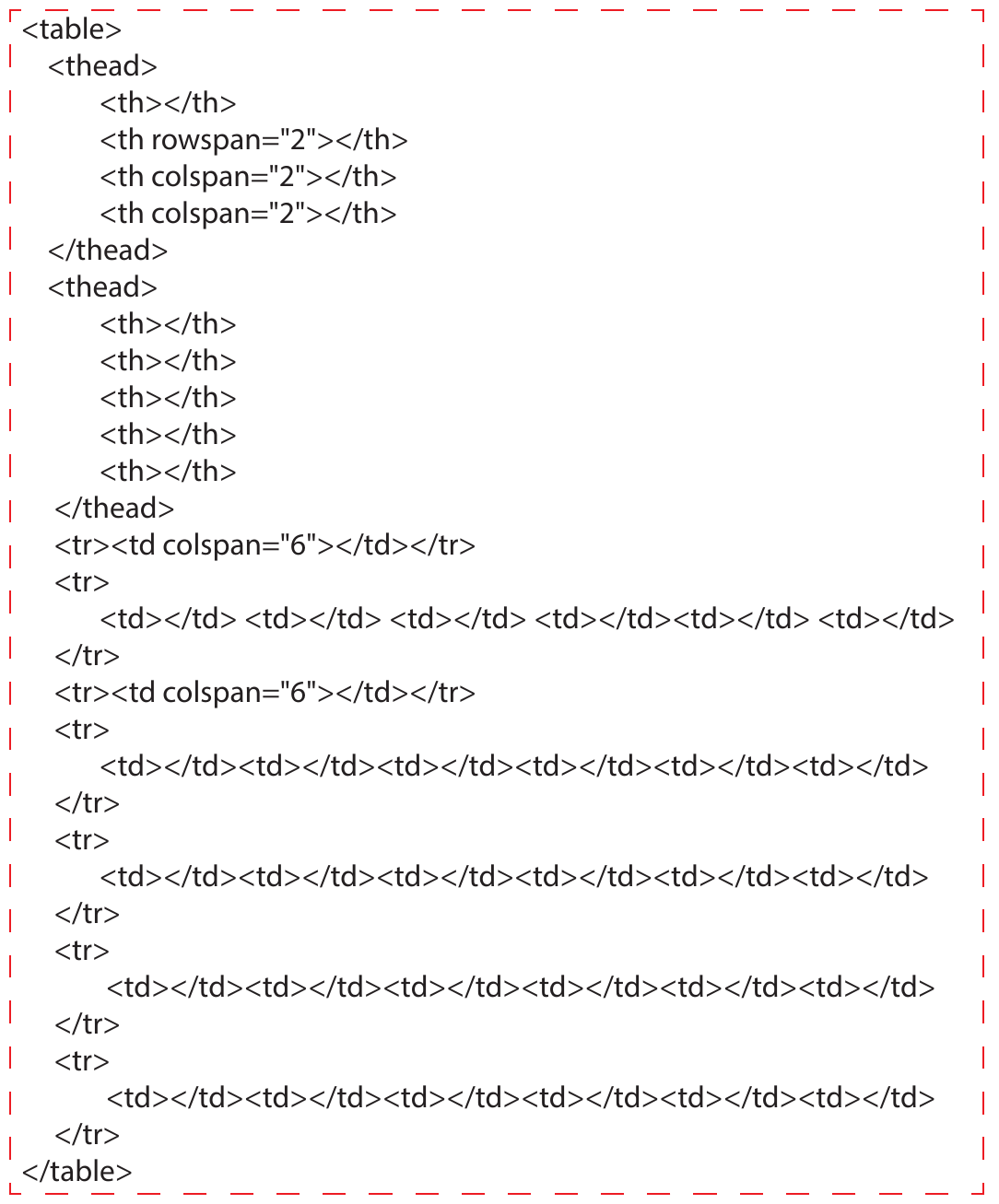}
         \caption{The generated HTML sequence after post-processing.}
         \label{fig:tsr_html}
     \end{subfigure}
    \caption{A sample of prediction result from the FinTabNet testing set.}
    \label{fig:prediction_sample}
\end{figure}

The experimental results regarding the structure-only TEDS and COCO metrics are shown in Tables~\ref{tab:scitsr_teds},~\ref{tab:fintabnet_teds},~\ref{tab:pubtables1m_teds},~\ref{tab:pubtabnet_teds} and~\ref{tab:results_ap}, which can demonstrate the superiority of the proposed solution. For the SciTSR dataset, the proposed TSRDet can improve the baseline Cascade R-CNN by 19.32\% regarding the structure-only TEDS, outperforming Deformable-DETR and Sparse R-CNN. When it comes to COCO metrics, the mAP of the proposed TSR is as good as Deformable-DETR, outperforming other benchmark models. Similarly, the proposed TSRDet can also outperform benchmark models regarding both COCO metrics and structure-only TEDS on the FinTabNet and PubTables1M datasets. It is worth mentioning that PubTabNet dataset does not provide original PDF files, making it hard to generate detection annotations. Therefore, the model performance reported in Table~\ref{tab:pubtabnet_teds} is calculated using the model trained with the PubTable1M dataset. Even though the PubTable1M and PubTabNet datasets have misalignments regarding the ground truth HTML sequences, the proposed method still shows competitive performance compared with other state-of-the-art methods, as shown in Table~\ref{tab:pubtabnet_teds}. Figure~\ref{fig:prediction_sample} shows a prediction sample and its generated HTML sequence after post-processing, demonstrating the capacities of the proposed solution. 

\begin{table}[!htpb]
\centering
\caption{Experimental results on PubTables1M dataset with structure-only TEDS score. Sim. means the tables without spanning cells and Com. represents the tables with spanning cells.}
\begin{tabular}{ c c c c} 
\hline
\multirow{2}{*}{Model} & \multicolumn{3}{c}{TEDS-struc.(\%)} \\
& Sim. & Com. & All  \\
\hline
 Cascade R-CNN & 82.73 & 85.21 & 83.78  \\
 Deformable-DETR & 97.54 & 93.14 & 95.73 \\
 Sparse R-CNN & 99.04 & 95.90 & 97.72 \\
 TSRDet(Ours) & $\mathbf{99.19}$ & $\mathbf{97.66}$ & $\mathbf{98.55}$ \\
\hline
\end{tabular}
\label{tab:pubtables1m_teds}
\end{table}

\begin{table}[!htpb]
\centering
\caption{Experimental results on PubTabNet validation set with structure-only TEDS score. Sim. means the tables without spanning cells and Com. represents the tables with spanning cells. The proposed model is trained with PubTable1M dataset, while the benchmark models are trained with PubTabNet dataset.}
\begin{tabular}{ c c c c} 
\hline
\multirow{2}{*}{Model} & \multicolumn{3}{c}{TEDS-struc.(\%)} \\
& Sim. & Com. & All  \\
\hline
EDD~\cite{zhong2020image} & 91.10 & 88.70 & 89.90\\
RobustTabNet~\cite{ma2023robust} & - & - & 97.00 \\
TSRNet~\cite{li2022table} & - & - & 95.64 \\
VAST~\cite{huang2023improving} & - & - & 97.23 \\
TableFormer~\cite{nassar2022tableformer} & 98.50 & 95.00 & 96.75 \\
MTL-TabNet~\cite{visapp23namly} & 99.05 & 96.66 & 97.88 \\
TSRDet(Ours) & 96.99 & 94.99 & 96.58 \\
\hline
\end{tabular}
\label{tab:pubtabnet_teds}
\end{table}

\begin{table*}[!htpb]
\caption{Experimental results with Average Precision.}
\begin{adjustbox}{width=\textwidth, center}
\begin{tabular}{ c c c c c c c c c c c c c c} 
\hline
Dataset & Model & $AP$ & $AP50$ & $AP75$ & $AP_s$ & $AP_m$ & $AP_l$ & Table & Column & Row & Spanning Cell & Projected Row Header & Column Header\\
 \hline
\multirow{4}{*}{SciTSR} & Cascade R-CNN & 93.89 & 95.27 & 94.80 & 95.81 & 93.89 & 92.96 & 98.96 & 98.63 & 96.33 & 88.58 & 83.80 & 97.01\\
 & Deformable-DETR & 96.28 & 97.39 & 97.01 & 96.75 & 96.55 & 96.07 & 98.96 & 98.63 & 97.26 & 93.84 & 90.86 & 98.15\\
  & Sparse R-CNN & 94.78 & 96.17 & 95.48 & 95.49 & 95.07 & 90.08 & 98.98 & 98.30 & 97.93 & 88.06 & 86.92 & 98.49\\
 & TSRDet(Ours) & 96.28 & 96.79 & 96.57 & 99.01 & 96.42 & 95.65 & 98.97 & 99.25 & 98.57 & 95.30 & 87.06 & 98.50 \\
\hline
\multirow{4}{*}{FinTabNet} & Cascade R-CNN & 95.23 & 97.53 & 96.90 & 87.32 & 95.31 & 93.08 & 99.00 & 96.69 & 96.96 & 84.43 & 96.63 & 97.64 \\
 & Deformable-DETR & 96.68 & 98.42 & 97.98 & 75.17 & 95.53 & 95.58 & 99.00 & 97.55 & 96.95 & 91.91 & 96.62 & 98.04\\
 & Sparse R-CNN & 96.38 & 98.37 & 97.69 & 62.11 & 96.22 & 95.86 & 99.01 & 97.79 & 97.84 & 88.39 & 97.29 & 97.97\\
 & TSRDet(Ours) & 97.50 & 98.33 & 98.09 & 91.60 & 97.40 & 97.15 & 99.01 & 98.83 & 97.99 & 94.62 & 96.61 & 97.93\\
\hline
\multirow{4}{*}{PubTables1M} & Cascade R-CNN  & 93.40 & 95.38 & 94.76 & 85.75 & 93.32 & 92.57 & 99.01 & 98.76 & 87.56 & 82.18 & 95.81 & 97.11\\
 & Deformable-DETR & 94.82 & 97.43 & 96.79 & 78.33 & 92.55 & 94.48 & 98.99 & 97.89 & 95.84 & 85.04 & 95.43 & 95.74\\
& Sparse R-CNN & 96.46 & 98.14 & 97.60 & 84.25 & 95.73 & 96.45 & 99.00 & 98.42 & 98.03 & 87.85 & 97.91 & 97.57\\
 & TSRDet(Ours) & 97.72 & 98.26 & 98.04 & 94.76 & 97.43 & 97.33 & 99.01 & 98.99 & 98.41 & 94.21 & 97.88 & 97.85\\
\hline
\end{tabular}
\end{adjustbox}
\label{tab:results_ap}
\end{table*}

\subsection{Ablation Study}
\label{sec:ablation_study}
In this section, we conduct experiments on the FinTabNet dataset to demonstrate the effectiveness of our applied methods, including using the proposed single-label detection formulation, tuning parameters of RPN, applying the deformable convolution and spatial attention. It is worth mentioning that tuning parameters of RPN includes increasing the number of proposals and ajusting the aspect rations. Since other studies have successfully applied the effectiveness of increasing the number of proposals, we applied it to both the Cascade R-CNN baseline and the proposed TSRDet, as discussed in section~\ref{sec:revisiting_region_proposal_generation} and~\ref{sec:expeirmental_results}. Therefore, we only discuss the the impact of adjusting the aspect ratios for tuning parameters of RPN in this section.

The experimental results are shown in Tables~\ref{tab:ablation_teds_fintabnet} and~\ref{tab:ablation_ap}, in which Asp\_Ratio Tuning, Single\_Label, DEFORM, and S\_Attn are shorts for applying aspect ratio tuning, single label formulation, deformable convolution, and spatial attention, respectively. Even though Cascade R-CNN baseline can reach 95.06\% regarding the mAP, its overall structure-only TEDS only reaches 82.70\%. After tuning the aspect ratios for the anchor generation, the structure-only TEDS is increased to 90.23\%, even though the mAP is only increased from 95.06\% to 95.54\%. Applying deformable convolution without other methods can improve the detection performance significantly but lead to a worse structure-only TEDS if we compare Ablation 1 and the Cascade R-CNN baseline.
Ablation 3 and Ablation 4 show that transforming the multi-label detection formulation into single-label formulation can significantly improve the performance, and also make deformable convolution improve the model performance. And applying both deformable convolution and spatial attention together can further improve the model performance from the results of Abliation 4, 5 and TSRDet, as shown in Table~\ref{tab:ablation_teds_fintabnet}. On the other hand, when it comes to detection metrics, applying deformable convolution always brings performance improvements from the results of Ablation 1 and Ablation 4, which can verify our analysis on the mismatch of detection metrics and cell-level metrics in section~\ref{sec:rethinking_detection_metrics_tsr_metrics}.

\begin{table}[!htpb]
\caption{Ablation study results on FinTabNet dataset with structure-only TEDS score. Asp\_Ratio Tuning, Single\_Label, DEFORM, and S\_Attn are shorts for applying aspect ratio tuning, single-label formulation, deformable convolution, and spatial attention.}
\centering
\begin{tabular}{ c c c c c c c c } 
\hline
\multirow{2}{*}{Model}& \multirow{2}{*}{Asp\_Ratio Tuning} & \multirow{2}{*}{Single\_Label} & \multirow{2}{*}{DEFORM} & \multirow{2}{*}{S\_Attn} & \multicolumn{3}{c}{TEDS-struc.(\%)}\\
& & & & & Sim. & Com. & All \\
\hline
Cascade R-CNN & & & & & 82.17 & 92.50 & 87.49  \\
Ablation 1 & & &\checkmark & & 81.45 & 87.11 & 84.35  \\
Ablation 2 & \checkmark & & & & 84.27 & 95.80 & 90.23  \\
Ablation 3 & \checkmark & \checkmark & & & 95.17 & 98.63 & 96.95  \\
Ablation 4 & \checkmark & \checkmark & \checkmark & &  96.44 & 99.14 & 97.83  \\
Ablation 5 & \checkmark & \checkmark & & \checkmark & 96.95 & 98.75 & 97.88  \\
TSRDet(Ours) & \checkmark & \checkmark & \checkmark & \checkmark & 99.08 & 99.02 & 99.05 \\
\hline
\end{tabular}
\label{tab:ablation_teds_fintabnet}
\end{table}

\begin{table*}[!htpb]
\caption{Ablation study results regarding Mean Average Precision. The model names are aligned with models in Table~\ref{tab:ablation_teds_fintabnet}.}
\begin{adjustbox}{width=\textwidth, center}
\begin{tabular}{ c c c c c c c c c c c c c c} 
\hline
Model & $AP$ & $AP50$ & $AP75$ & $AP_s$ & $AP_m$ & $AP_l$ & Table & Column & Row & Spanning Cell & Projected Row Header & Column Header\\
\hline
Cascade R-CNN  & 95.23 & 97.53 & 96.90 & 87.32 & 95.31 & 93.08 & 99.00 & 96.69 & 96.96 & 84.43 & 96.63 & 97.64 \\
Ablation 1 & 97.22 & 98.03 & 97.90 & 90.11 & 96.72 & 96.76 & 99.00 & 98.95 & 96.16 & 94.98 & 96.01 & 98.19 \\
Ablation 2 & 95.54 & 97.54 & 96.91 & 87.43 & 95.79 & 94.04 & 99.00 & 97.04 & 97.64 & 84.84 & 96.67 & 98.02 \\
Ablation 3 & 95.51 & 97.56 & 96.94 & 88.43 & 95.52 & 93.76 & 99.00 & 97.31 & 97.87 & 84.74 & 96.82 & 97.28 \\
Ablation 4 & 97.83 & 98.37 & 98.13 & 91.91 & 97.65 & 97.58 & 99.00 & 98.96 & 98.33 & 95.78 & 96.98 & 97.93 \\
Ablation 5 & 96.97 & 97.84 & 97.58 & 90.32 & 96.88 & 96.21 & 99.00 & 98.83 & 98.03 & 91.97 & 96.58 & 97.37 \\
TSRDet(Ours) & 97.50 & 98.33 & 98.09 & 91.60 & 97.40 & 97.15 & 99.01 & 98.83 & 97.99 & 94.62 & 96.61 & 97.93\\

\hline
\end{tabular}
\end{adjustbox}
\label{tab:ablation_ap}
\end{table*}

\subsection{Observations and Analysis}
\label{sec:obvservations_and_analysis}
Sections~\ref{sec:expeirmental_results} and~\ref{sec:ablation_study} have demonstrated the effectiveness of our proposed methods. In this section, we further discuss some observations from the experimental results and how these observations verify our analysis in section~\ref{sec:rethinking_detection_based_models}. 

\subsubsection{Multi-label Detection}
As discussed in sections~\ref{sec:prelimiaries} and~\ref{sec:revisiting_problem_formulations}, multi-label detection tasks are difficult for two-stage detection models, but transformer-based detection models with learnable proposals can deal with multi-label detection tasks. Besides, the problem formulation of PubTables1M is a multi-label task, making it difficult for two-stage detection models. The experimental results from sections~\ref{sec:expeirmental_results} can demonstrate our analysis. For example, as shown in Table~\ref{tab:scitsr_teds}, the performance of Deformable-DETR and Sparse R-CNN are 97.81\% and 98.13\% regarding the structure-only TEDS, which are very close to the performance of proposed TSRDet (98.41\%) and far better than the Cascade R-CNN baseline (79.09\%). Notably, as mentioned in section~\ref{sec:experiments}, all the models, except our proposed TSRDet, are using the multi-label detection setting. Therefore, two trasnformer-based detection models show promising results in the multi-label detection setting. Similarly, the experiments on the FinTabNet and PubTables1M datasets also show similar results. For example, the structure-only TEDS performance of Sparse R-CNN, TSRDet, and Cascade R-CNN baseline are 97.72\%, 98.55\%, and 83.78\% on the PubTables1M dataset, 98.13, 99.05, and 87.49 on the FinTabNet dataset.

\subsubsection{The Misalignment of Metrics}
\begin{figure}
     \centering
     \begin{subfigure}[b]{\columnwidth}
         \centering
         \includegraphics[width=0.8\columnwidth]{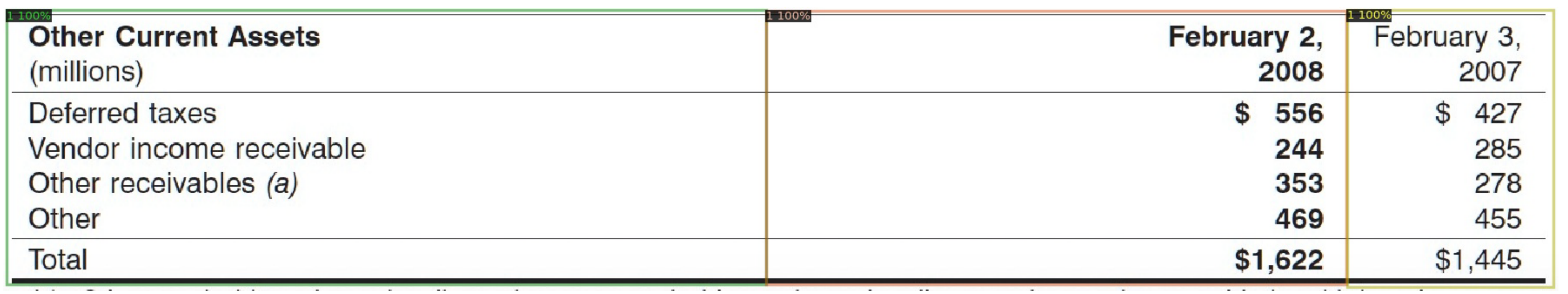}
         \caption{A sample prediction result of the Ablation 1 model. The mAP and structure-only TEDS are 97.22 and 84.35, respectively. We only include the columns' predictions for simplicity.}
         \label{fig:misalignment_ablation1}
     \end{subfigure}
     \hfill
     \begin{subfigure}[b]{\columnwidth}
         \centering
         \includegraphics[width=0.8\columnwidth]{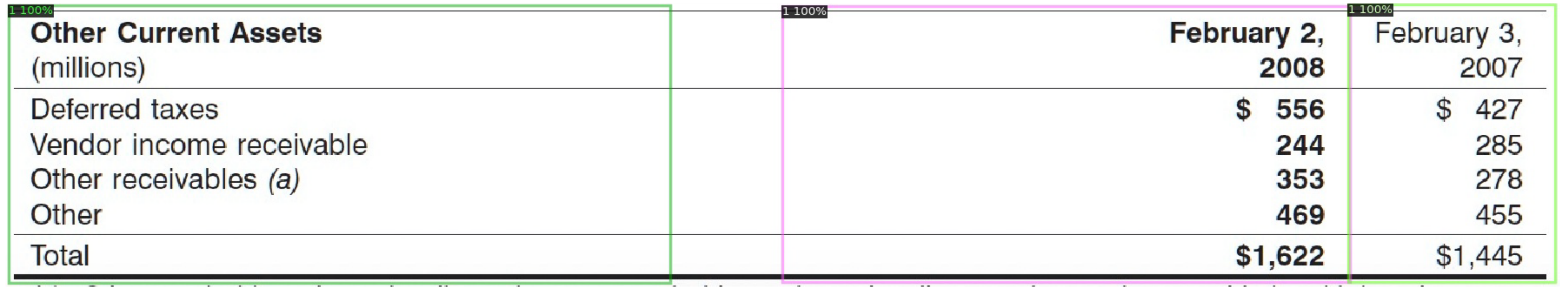}
         \caption{A sample prediction result of the Ablation 3 model. The mAP and structure-only TEDS are 95.51 and 96.95, respectively. We only include the columns' predictions for simplicity.}
         \label{fig:misalignment_ablation3}
     \end{subfigure}
    \caption{Comparison of results from Ablation1 and Ablation3 models. Even though Ablation 1 can achieve better detection performance, its performance regarding structure-only TEDS is much lower than that of Ablation 3 model.}
    \label{fig:metrics_misalignment_explaination}
\end{figure}

The experimental results in sections~\ref{sec:expeirmental_results} and~\ref{sec:ablation_study} show the misalignment of COCO and TEDS metrics many times. For example, in Table~\ref{tab:results_ap}, both the Deformable-DETR and our proposed TSRDet can reach 96.28\% regarding mAP on the SciTSR dataset, which is better than that of Sparse R-CNN. However, when it comes to structure-only TEDS, as shown in Table~\ref{tab:scitsr_teds}, both TSRDet and Sparse R-CNN can perform better than Deformable-DETR. Similar results also appear in the experiments on the FinTabNet dataset. As shown in Table~\ref{tab:results_ap}, the mAP of Sparse-RCNN and Deformable-DETR are 96.38\% and 96.68\%, while their structure-only TEDS are 97.81\% and 98.13\%. More similar results can be found in the results of the ablation study, such as Ablation 1 and Ablation 3, as shown in Tables~\ref{tab:ablation_teds_fintabnet} and~\ref{tab:ablation_ap}. To further verify our discussion in section~\ref{sec:rethinking_detection_metrics_tsr_metrics}, we show the prediction results of Ablation 1 and 3 in Figure~\ref{fig:metrics_misalignment_explaination}. As discussed in section~\ref{sec:rethinking_detection_metrics_tsr_metrics}, COCO metrics are relied on IoU scores, while TEDS is not. Therefore, as shown in Figure~\ref{fig:misalignment_ablation1}, Ablation 1 with deformable convolution can better fit the extra white areas to improve the detection performance, but it cannot improve the TEDS compared with Ablation 3 whose result is shown in Figure~\ref{fig:misalignment_ablation3}. It is worth mentioning that the ground truth of the sample in Figure~\ref{fig:metrics_misalignment_explaination} has been shown in Figure~\ref{fig:metrics_misalignment_sample}.

\subsubsection{Deformable Convolution and Spatial Attention}
As discussed in sections~\ref{sec:rethinking_detection_metrics_tsr_metrics} and~\ref{sec:rethinking_feature_generation}, both generating good local features and building long-range dependencies are essential for a detection-based TSR model, and deformable convolution can improve the local feature generation but has the risk leading to the over-optimization to the detection performance. The ablation study's experimental results can somewhat demonstrate our analysis. Considering the performance of Ablation 1 with deformable convolution in Tables~\ref{tab:ablation_teds_fintabnet} and~\ref{tab:ablation_ap}, its TEDS is 84.35\%, lower than the Cascade R-CNN baseline (87.49\%), but its mAP is improved from 95.23\% to 97.22\%. These results not only show the misalignment of COCO and TEDS metrics but also demonstrate that merely improving local features can make the model fit empty spaces better, as shown in Figures~\ref{fig:misalignment_ablation1}, but does not help alleviate the multi-label detection issue. Therefore, deformable convolution needs to be applied with other methods. On the other hand, our proposed Spatial Attention Module can improve the mAP and structure-only TEDS simultaneously if we compare the performance of Ablation 3 and 4, and also can be used with deformable convolution together to improve the structure-only TEDS further, as shown in Figure~\ref{tab:ablation_teds_fintabnet}, demonstrating the effectiveness of building long-range dependencies.  

\subsubsection{Other Observations}
\label{sec:other_observations}
Besides the observations discussed in previous sections, the experimental results also show other phenomena that can be helpful in our model design. One observation is that Cascade R-CNN has better detection performance on small objects than Sparse R-CNN. For example, on the FinTabNet dataset, $APs$ of Sparse R-CNN is only 62.11\%, while the Cascade R-CNN baseline and our proposed TSRDet reach 87.32\% and 91.60\%. This phenomenon might be caused by their methods of generating regional proposals. As discussed in section~\ref{sec:prelimiaries}, Cascade R-CNN uses RPN to generate regional proposals, which regress and classify anchor boxes, and the anchor boxes are generated by sliding the pre-defined boxes with different aspect ratios and sizes on the feature map of multiple scales. Therefore, Cascade R-CNN uses a dense proposal generation method~\cite{sun2021sparse} with more region proposals, meaning that Cascade R-CNN can use the parameters of RPN to generate more high-quality small region proposals. By contrast, Sparse R-CNN uses sparse learnable regional proposals to replace dense proposals generated by the RPN, which can avoid parameter tuning of RPN but limit its performance on small objects. 
Another interesting observation is that the baseline Cascade R-CNN can work better on complex tables than simple tables, which is very different from other benchmark models. This phenomenon is caused by the fact that the spanning cells in complex tables are usually in the Column Row Headers, which can alleviate the multi-label detection issue. For example, Figures~\ref{fig:definition_6classes} and Figures~\ref{fig:definition_6classes_2} show two samples from PubTables1M dataset, in which the former is a complex table and the latter is a simple table. Because of the existence of Spanning Cells in  Figure~\ref{fig:definition_6classes}, the Column Header does not share its bounding box with any rows, which avoids multi-label detection. By contrast, the sample in Figure~\ref{fig:definition_6classes_2} does not contain any Spanning Cell, making its Column Header share its bounding box with a Row, which is the challenging multi-label detection to Cascade R-CNN. As comparisons, Deforamble-DETR and Sparse R-CNN can deal with multi-label detection, and their performance on simple tables is better than complex tables regarding the structure-only TEDS, as shown in Tables~\ref{tab:scitsr_teds},~\ref{tab:fintabnet_teds}and~\ref{tab:pubtables1m_teds}.

\section{Conclusion and Future Work}
\label{sec:conclusion}
In this study, we first revisit existing detection-based TSR solutions and analyze the critical design aspects for a successful detection-based TSR model, including the problem formulation, the characteristics of detection models, and the characteristics of TSR tasks. Our analysis can be a guideline for improving the performance of a detection-based model. To demonstrate our analysis and findings, we propose TSRDet by applying simple methods to tailor the Cascade R-CNN, which can outperform different types of state-of-the-art models, including image-to-sequence and graph-based models. Even though we only applied very simple methods to a tow-stage detection model, there should be other methods to further improve the model based on our analysis. For example, vision transformers can be considered to build long-range dependencies. Transformer-based detection models, such as Sparse R-CNN, can also be considered as base models with the benefits of dealing with multi-label detection tasks and learnable proposals. Besides, since the proposed method is detection-based and focuses on well-formatted, visually rich documents, one major limitation is that it may fail to deal with irregular tables, such as rotated and distorted tables. Integrating instance segmentation with detection models can be another direction to deal with irregular tables because instance segmentation can handle irregular shapes and be guided by bounding boxes.

\section*{Acknowledgements}
 This work was supported in part by Mathematics of Information Technology and Complex Systems (Mitacs) Accelerate Program and Lytica Inc.

 \bibliographystyle{elsarticle-num} 
 \bibliography{reference.bib}





\end{document}